\documentclass[10pt,journal,compsoc]{IEEEtran}

\ifCLASSOPTIONcompsoc
  \usepackage[nocompress]{cite}
  \usepackage{cite}
  \usepackage{graphicx}
  \usepackage{bm}
  \usepackage{epsfig}
  \usepackage{amsmath}
  \usepackage{array}
  \usepackage{multirow}
  \usepackage{color}
  \usepackage{mathrsfs}
  \usepackage{amsfonts}
  \usepackage{caption}
  \usepackage{subfigure}
  \usepackage{amsthm}
  \usepackage{amssymb}
\else
  \usepackage{cite}
\fi
\ifCLASSINFOpdf
\else
\fi
\begin{document}
%
\title{Multi-task Learning with Coarse Priors for Robust Part-aware Person Re-identification}
%
%
%
%

\author{Changxing Ding*,~\IEEEmembership{Member,~IEEE,}
        Kan Wang*,
        Pengfei Wang,
        and Dacheng Tao,~\IEEEmembership{Fellow,~IEEE}

\thanks{C. Ding, K. Wang, and P. Wang are with the School of Electronic and Information Engineering,
South China University of Technology, 381 Wushan Road, Tianhe District, Guangzhou 510000, P.R. China.
(e-mail: chxding@scut.edu.cn; eekan.wang@mail.scut.edu.cn; mswangpengfei@mail.scut.edu.cn).}
\thanks{D. Tao is with the School of Computer Science, in the Faculty of Engineering, at The University of Sydney,
6 Cleveland St, Darlington, NSW 2008, Australia (email: dacheng.tao@sydney.edu.au).}
\thanks{* indicates equal contribution.}
}

\IEEEtitleabstractindextext{%
\begin{abstract}
Part-level representations are important for robust person re-identification (ReID),
but in practice feature quality suffers due to the body part misalignment problem.
In this paper, we present a robust, compact, and easy-to-use method called the Multi-task Part-aware Network (MPN),
which is designed to extract semantically aligned part-level features from pedestrian images.
MPN solves the body part misalignment problem via multi-task learning (MTL) in the training stage.
More specifically, it builds one main task (MT) and one auxiliary task (AT) for each body part on the top of the same backbone model.
The ATs are equipped with a coarse prior of the body part locations for training images.
ATs then transfer the concept of the body parts to the MTs via optimizing the MT parameters to identify part-relevant channels from the backbone model.
Concept transfer is accomplished by means of two novel alignment strategies: namely, parameter space alignment via hard parameter sharing and feature space alignment in a class-wise manner.
With the aid of the learned high-quality parameters,
MTs can independently extract semantically aligned part-level features from relevant channels in the testing stage.
MPN has three key advantages: 1) it does not need to conduct body part detection in the inference stage; 2) its model is very compact and efficient for both training and testing;
3) in the training stage, it requires only coarse priors of body part locations, which are easy to obtain.
Systematic experiments on four large-scale ReID databases demonstrate that MPN consistently outperforms state-of-the-art approaches by significant margins.
\end{abstract}

}

\maketitle

\IEEEdisplaynontitleabstractindextext

%
\IEEEpeerreviewmaketitle

\IEEEraisesectionheading{\section{Introduction}\label{sec:introduction}}

%
%
%
%

\IEEEPARstart{P}{erson} re-identification (ReID) is a critical component of modern surveillance systems.
The process is aimed at spotting a person of interest, e.g. a missing child or a suspect, across disjoint camera views distributed at different physical locations.
Due to the widespread deployment of visual surveillance networks, ReID has recently attracted increasing attention from both academia and industry.
Despite this, however, ReID remains a challenging problem; this is largely caused by the dramatic variations in intra-personal appearance and high inter-personal similarity~\cite{karanam2019systematic,chen2017person,li2019unsupervised,li2017person,lisanti2014person,bai2017scalable}.
Accordingly, to enhance the discriminative power of ReID models, a large proportion of the recent literature has explored the learning of part-level representations~\cite{2wang2018learning,3zhao2017spindle,4suh2018part,8zhang2018densely,29xu2018attention,32Part-based2017tip,
33zhang2018person,36Cheng2016Person,39li2017learning,40zhao2017deeply,60fu2018horizontal},
which incorporate more fine-grained features and reduce the overfitting risk of deep models~\cite{6yao2017deep,5sun2017beyond}.

\begin{figure}
\centering
\includegraphics[width=1.0\linewidth]{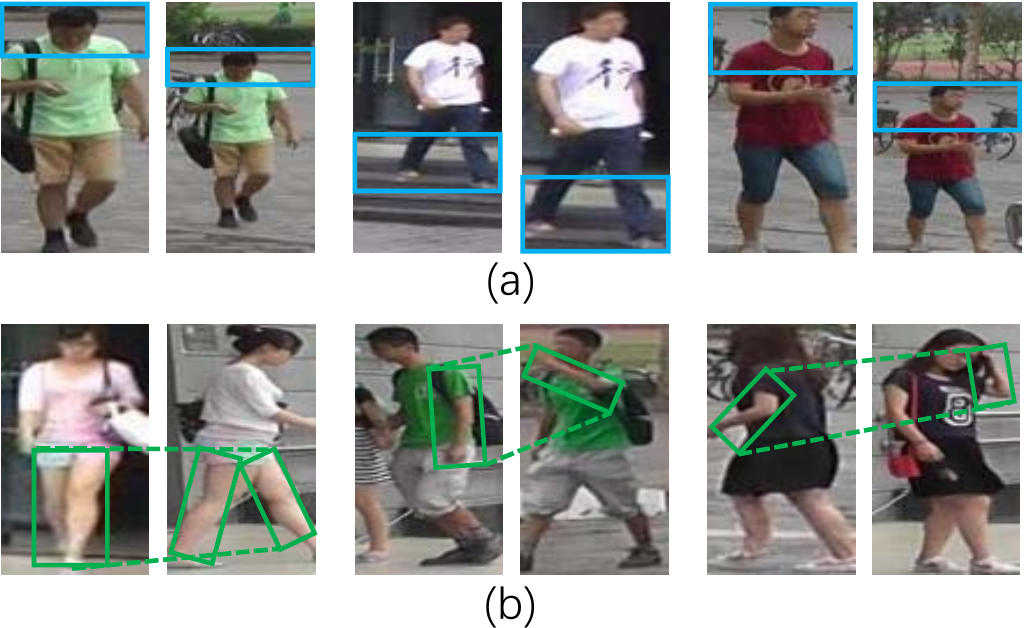}
\caption{The spatial misalignment of body parts across images is common in pedestrian images.
(a) Misalignment caused by errors in pedestrian detection.
Both the position and scale of the body parts change, as indicated by the blue rectangles.
(b) Misalignment due to the movement of flexible body parts, such as arms and legs.
Both the position and shape of these body parts vary, as indicated by the green rectangles.
(Best viewed in color.)
}
\label{fig:partMisalignment}
\end{figure}

However, the extraction of high-quality part-level representations is difficult.
This is because human body parts are often not semantically aligned across images,
meaning that the same spatial position across two images may not correspond to the same body part.
As illustrated in Fig.~\ref{fig:partMisalignment}, one reason is that pedestrian detection is still challenging~\cite{zhang2017s3fd,81faster,chi2019pedhunter}.
The error in pedestrian detection causes both the position and scale of body parts to vary dramatically in images.
Moreover, some body parts, such as arms and legs, are inherently flexible,
meaning that their position and shape will change even in cases where the pedestrian detection algorithm works perfectly.

Since the positions of these body parts are variable, one intuitive solution is to detect body parts in the spatial dimension before part-level feature extraction occurs;
most existing works adopt this approach~\cite{6yao2017deep,38li2018harmonious,40zhao2017deeply,5sun2017beyond}.
However, body part detection is inherently challenging:
interference such as severe image blur, background clutter, and occlusions
may cause failures in part detection, thereby degrading the quality of the part-level representations.
There have been recent attempts to bypass part detection at the inference stage using teacher-student style training strategies~\cite{8zhang2018densely,ren2018factorized}.
The teacher model employs prior information regarding body part locations to guide a separate student model in extracting semantically aligned part-level features from the original image.
However, there are two disadvantages to this approach. Firstly, the architecture of the model used for training is complex.
Secondly, only the feature space is constrained in a sample-wise manner,
with the result that the performance of the student model is sensitive to the robustness of the teacher model.

It is therefore reasonable to seek a robust, compact, and easy-to-use method capable of learning semantically aligned part-level representations.
To this end, we propose a Multi-task Part-aware Network (MPN) that only slightly increases the time and space complexities of a very basic part-based model~\cite{5sun2017beyond} for both training and testing.
Unlike existing works, MPN extracts part-specific information from a deep backbone model
by \textit{explicitly} regularizing the model parameters to select part-relevant channels via the introduction of inductive bias with multi-task learning (MTL).
Global max-pooling on selected channels results in translation- and scale-invariant body part features being obtained.
Moreover, as the selected channels for each body part are fixed after training, MPN is naturally robust to various forms of interference such as image blur and background clutter.

One primary contribution of MPN is the way it explicitly learns part-relevant channels.
The first challenge is that there is no universally recognized definition of the area for each body part.
Moreover, a plain network cannot learn the concept of `body parts' without any priors.
In this paper, we define the body part regions in training images using a method that is coarse but robust and easy to use.
In the training stage, MPN includes one main task (MT) and one auxiliary task (AT) for each body part, both of which are built on the same backbone model.
Both tasks select and combine body part relevant channels to construct the respective part-level representations for person classification.
Their main difference lies in the input feature maps: the inputs of MT are the original feature maps produced by the backbone model,
while those for AT are the cropped feature maps according to the coarse part location priors.
By sharing all parameters (except for their respective classifiers) of MT and AT for each respective part,
MT is regularized to enable the choosing of part-relevant channels.
Another noticeable advantage of this strategy is that the obtained model architecture for training is extremely compact.

Implementing the above simple strategy enables MPN to achieve state-of-the-art performance.
However, due to the difference between the input feature maps, the sharing parameter alone cannot ensure alignment of the MT and AT feature spaces.
Accordingly, to eliminate this discrepancy, we further introduce a novel constraint between MT and AT in the feature space,
which is applied in a class-wise rather than a sample-wise manner~\cite{8zhang2018densely,ren2018factorized}.
Briefly, we compute the mean representation of each identity in one batch for MT and AT respectively, and then penalize their cosine distance.
The motivation behind this is that the prior location of each body part employed by AT is coarse, which degrades the quality of the feature vectors obtained by AT;
therefore, a feature space constraint at the class level will statistically be more robust than one at the sample level~\cite{8zhang2018densely,ren2018factorized}.

In the inference stage, all ATs are abandoned and part-level features obtained by MTs are used for ReID.
To demonstrate the efficacy of MPN, we conduct extensive experiments on four large-scale benchmark datasets:
Market-1501~\cite{27zheng2015scalable}, DukeMTMC-ReID~\cite{59zheng2017unlabeled}, CUHK03~\cite{33li2014deepreid}, and MSMT17~\cite{80PTGAN}.
The results show that our simple MPN model consistently and significantly outperforms existing approaches
with the further advantages of being compact and easy to use.

The remainder of the paper is organized as follows. Related works on ReID and MTL are briefly reviewed in Section 2.
The MPN model structure and training scheme are described in Section 3.
ReID during inference using MPN is introduced in Section 4.
Detailed experiments and their analysis are presented in Section 5. We conclude in Section 6.

\section{Related Works}
A number of effective approaches have been proposed for ReID~\cite{zheng2016person}.
In particular, part-based models have been shown to be effective and have become popular~\cite{6yao2017deep,5sun2017beyond,8zhang2018densely}.
We therefore review the literature on (i) part-based ReID models and (ii) MTL methods for ReID.

\subsection{Part-based ReID Models}
While part-based models are powerful, they suffer from the problem of the semantic misalignment of body parts~\cite{6yao2017deep,5sun2017beyond}.
Existing approaches to this problem can be divided into three categories:
(i) methods that extract multi-scale features (MSF) to address the body part misalignment problem;
(ii) methods that detect body parts in the spatial dimension before part-level feature extraction is performed;
and (iii) methods that guide the deep model to learn semantically aligned part-level features via teacher-student training schemes.

MSF-based methods extract part-level features from multi-scale image patches~\cite{2wang2018learning,60fu2018horizontal,102zheng2019pyramidal}.
As the patch size increases, the extracted features are less affected by body part misalignment
but at the cost of reduced discriminative power.
Multi-scale patch features are concatenated as the image representation.
However, the dimension of the final representation is high, and the body part misalignment problem is only partly solved.

Body part detection-based methods first detect body parts in the spatial dimension before part-level feature extraction is performed.
Most existing works fall into this category.
According to the way in which body parts are detected, approaches in this category can be further classified into
(i) outside tools-based methods, (ii) spatial attention-based methods, and (iii) unsupervised methods.
Outside tools-based methods rely on outside tools (e.g., pose estimation models~\cite{3zhao2017spindle,11su2017pose,29xu2018attention,69Pose}
and human parsing models~\cite{kalayeh2018human}), to provide body part locations during both training and testing.
Notable downsides to this approach include the extra computational cost and the low reliability of the outside tools.
Spatial attention-based methods can overcome the above problem by inferring the location of body parts directly from feature maps produced by ReID networks~\cite{38li2018harmonious,40zhao2017deeply,42liu2017hydraplus,34macus2018}.
The output of the attention modules can take the form of either rigid bounding boxes~\cite{38li2018harmonious} or soft spatial masks~\cite{40zhao2017deeply,42liu2017hydraplus,5sun2017beyond} that represent the location of body parts.
Attention module parameters are optimized together with the entire ReID network using supervision signals for ReID only,
meaning that the parameters of the attention modules lack a direct constraint.
Unsupervised methods adopt hand-crafted approaches to locating body parts based on the feature maps of each image~\cite{6yao2017deep,zhang2017alignedreid}.
For example, Yao \emph{et al.}~\cite{6yao2017deep} proposed using K-means clustering to cluster channels based on the locations of maximum response,
with individual average pooling of the selected channels for each cluster indicating the location of a body part.

Despite these efforts, however, body part detection-based methods still face major challenges,
because body part detection can fail when the image contains interference such as severe image blur, background clutter, and significant occlusion.

The third category of methods bypass body part detection during inference~\cite{8zhang2018densely,ren2018factorized}.
In the training stage, these methods adopt complex model architectures with teacher-student-style training strategies.
As the teacher model is equipped with prior information regarding body part locations,
it can extract semantically aligned part-level features.
Moreover, through alignment with the teacher model in the feature space, a separate student model without any body part priors is guided to produce similar features.
Existing works guide in a sample-wise manner~\cite{8zhang2018densely,ren2018factorized}; therefore, the quality of guidance is vital.
To provide precise guidance, Zhang~\emph{et al.}~\cite{8zhang2018densely} used a 3D alignment tool to achieve pixel-level semantic alignment of body parts.
However, the performance of the outside tool used was restricted by training image quality
(e.g., severe image blur), with the result that the obtained body part location priors may not be robust.
Therefore, sample-wise guidance between the teacher and student models may be suboptimal for ReID.

Compared with existing works, our proposed approach not only bypasses body part detection during inference,
but also has the advantages of compactness, robustness, and ease-of-use both during training and testing.
In particular, we here solve the body part misalignment problem from a novel perspective: in short, we
explicitly select part-relevant channels from the backbone model by introducing inductive bias with auxiliary tasks.

\subsection{MTL Methods for ReID}
MTL is a commonly used strategy that simultaneously optimizes multiple relevant tasks.
These relevant tasks introduce inductive bias, which improves the generalization ability of the main task.
Therefore, MTL has been successfully applied in many computer vision tasks.
Here we focus on MTL-based approaches for ReID.
For a more comprehensive summary of MTL, we direct readers to~\cite{ruder2017overview}.

Existing MTL approaches for ReID can be divided into four categories.
First, many works combine loss functions for image classification and metric learning to improve the quality of learned pedestrian representations~\cite{25chen2017multi,34macus2018}.
Second, other models conduct person ReID and attribute recognition jointly,
since these tasks are closely related~\cite{su2017multi,mclaughlin2016person,wang2018transferable}.
Third, some recent part-based approaches have generated feature maps for each body part from the output of the same backbone model,
regarding ReID based on each body part as an independent task~\cite{6yao2017deep,5sun2017beyond,2wang2018learning}.
Fourth, body part detection and part-based ReID were integrated into a single model as two parallel tasks during training in~\cite{wang2019cdpm}.
In the testing phase, these two tasks run sequentially for ReID.

In this paper, we employ MTL for a new purpose. Briefly, MTL regularizes the model parameters to select channels relevant to each respective body part,
such that the subsequently extracted part-level features are semantically aligned.
Moreover, compared with~\cite{2wang2018learning,8zhang2018densely,wang2019cdpm},
our MTL-based approach is very compact due to its use of hard parameter sharing.

\section {Multi-task Part-Aware Network}
We first introduce the motivation and problem formulation of MPN before
presenting the MPN framework and describing each of its key components:
namely, the coarse priors of part locations for training images, the part-relevant channel selection via MTL, and the class-wise feature space alignment (FSA) between the two tasks.

\begin{figure}
\centering
\includegraphics[width=0.9\linewidth]{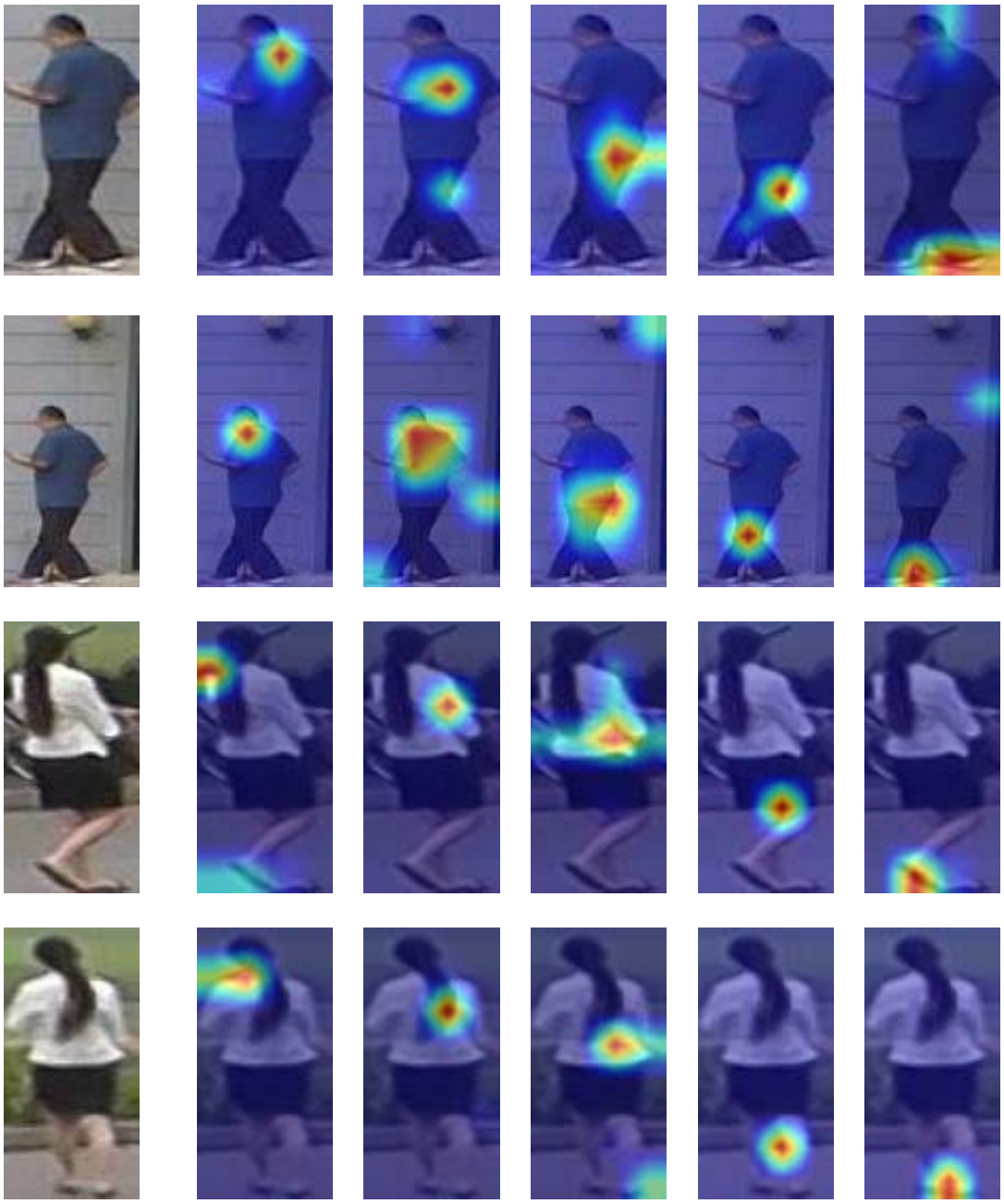}
\caption{Examples of per-channel responses (heatmaps) on the human body.
The first column presents the original images.
Each of the other five columns represents responses on one representative channel, respectively.
The channels are selected from the last convolutional layer of the ResNet-50 model. Red denotes stronger activation.
The figure illustrates that there are correspondences between each body part and different channels.
}
\label{fig:channelResponse}
\end{figure}

\subsection{Problem Formulation}
Recent works have shown that different channels of a ReID network activate local responses at their corresponding body parts~\cite{6yao2017deep,14zhang2018occluded}.
In other words, there are correspondences between the channels and body parts, as shown in Fig.~\ref{fig:channelResponse}.
This property has been utilized to detect body parts in the spatial dimension~\cite{6yao2017deep,wang2019cdpm}.

In this paper, we explore this property from a more straightforward perspective.
During training, we train a light-weight module, i.e., one $1\times1$ convolutional (Conv) layer,
to select and combine relevant channels for each individual body part from the output feature maps of a deep backbone model.
Global max-pooling (GMP) on each of the produced channels results in translation- and scale-invariant body part features.
The parameters of the $1\times1$ Conv layers are fixed during the inference stage;
as a result, part-level features can be robustly extracted without the need for body part detection for each image.

The problem is then reformulated for the identification of the relevant channels for each body part.
Unfortunately, a plain deep model cannot automatically learn the concept of body parts without proper guidance.
Accordingly, in this section, we propose MPN to solve this problem via MTL.

\begin{figure*}
\centerline{\includegraphics[width=1.0\textwidth]{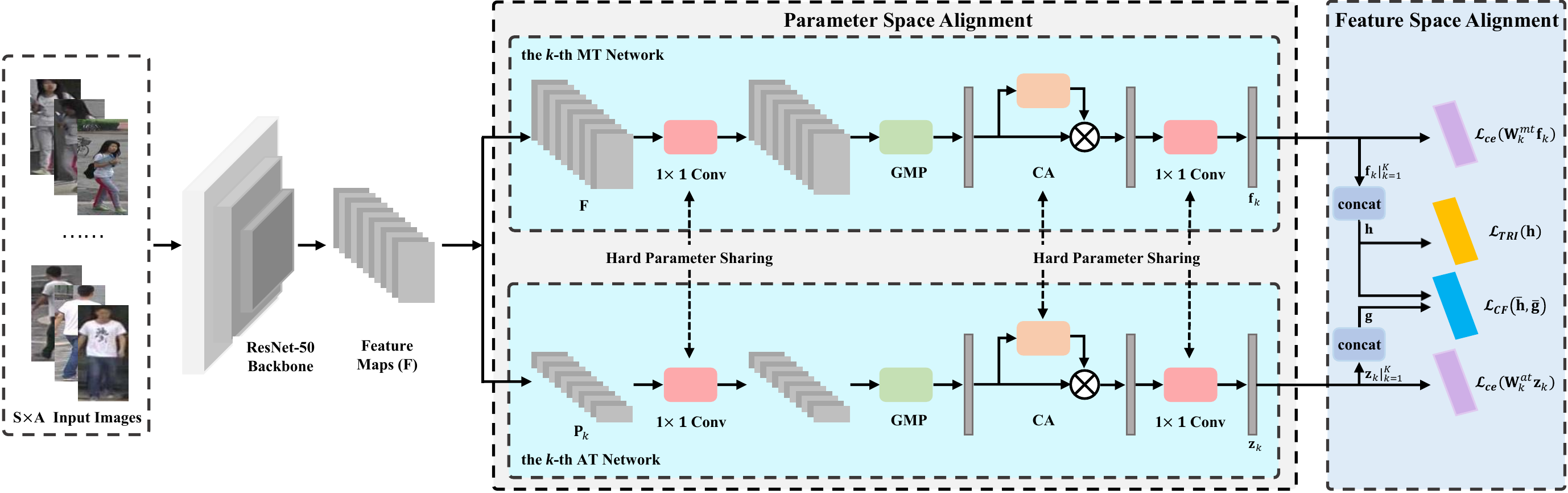}}
\caption{
Model architecture of MPN in the training stage. Based on the ResNet-50 backbone model, MPN builds two tasks for each of the $K$ body parts:
one main task (MT) and one auxiliary task (AT). For simplicity, only one MT-AT pair is shown in this figure.
The $K$ ATs are equipped with the coarse prior of body part locations for the training images,
which provides inductive bias assisting the MTs to select and combine the relevant channels for each body part.
Inductive bias is transferred via two key operations: parameter and feature space alignments between each MT-AT pair.
The selected part-relevant channels are processed to obtain part-level representations.
In the inference stage, all ATs are removed to leave only the MTs to extract image representations.
}
\label{fig:framework}
\end{figure*}

As illustrated in Fig.~\ref{fig:framework}, MPN in the training stage includes two tasks for each of $K$ body parts:
one main task (MT) and one auxiliary task (AT).
Both tasks are optimized for ReID purposes, i.e. person classification.
The $K$ ATs are equipped with the coarse prior of the body part locations in the training images;
as a result, they can provide inductive bias assisting the MTs to select the relevant channels for each body part.
In the inference stage, ATs are removed, and only the MTs are used to extract part-level representations.
We next introduce each of the key components of MPN.

\subsection{Coarse Prior of Body Part Locations}
\label{sec:part_prior}
The prior of body part locations for training images provides the network with the concept of body parts.
Body parts can be represented as size-fixed strips~\cite{2wang2018learning,102zheng2019pyramidal},
size-varied bounding boxes~\cite{6yao2017deep}, or even a group of pixels of irregular shape~\cite{8zhang2018densely}.
We here adopt the strip-based representation, which is coarse but robust. The prior is generated based on two existing tools:
one for human parsing~\cite{1Liao2016understand} and another for human segmentation~\cite{12song2018mask}.

\begin{figure}
\centering
\includegraphics[width=1.0\linewidth]{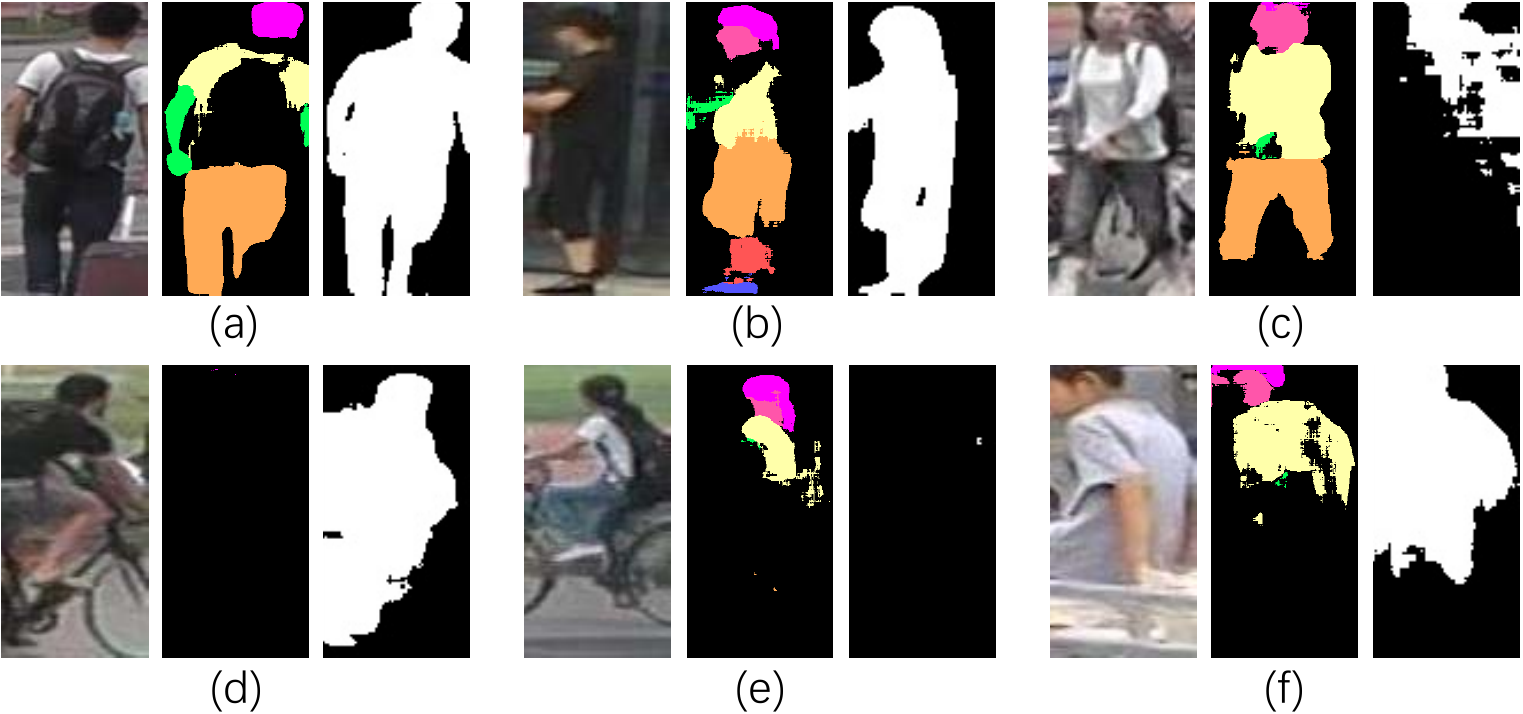}
\caption{
The three images in each group show the pedestrian image, human parsing result by~\cite{1Liao2016understand}, and human segmentation result by~\cite{12song2018mask}, respectively.
(a) The tool in~\cite{1Liao2016understand} may ignore discriminative accessories, e.g., a backpack.
(b) It may also neglect undefined body parts, e.g., the neck.
(c-d) Each tool may fail if the image quality is low.
(e) A situation in which both tools have failed to work.
(f) Segmentation results when there is a severe part missing problem.}
\label{fig:mask}
\end{figure}

As explained in Fig.~\ref{fig:mask}(a-b),
the former tool in~\cite{1Liao2016understand} segments and distinguishes a set of pre-defined body parts,
but ignores discriminative accessories (e.g., backpacks) and undefined body parts (e.g., necks).
The latter tool in~\cite{12song2018mask} segments the human body with accessories as a whole, thereby losing part-specific information.
As shown in Fig.~\ref{fig:mask}(c-d), each of them may fail;
however, the chance that they both fail for the same image (e.g., Fig.~\ref{fig:mask}(e)) is small. Therefore, the two tools are complementary.

\begin{figure}
\centering
\includegraphics[width=1.0\linewidth]{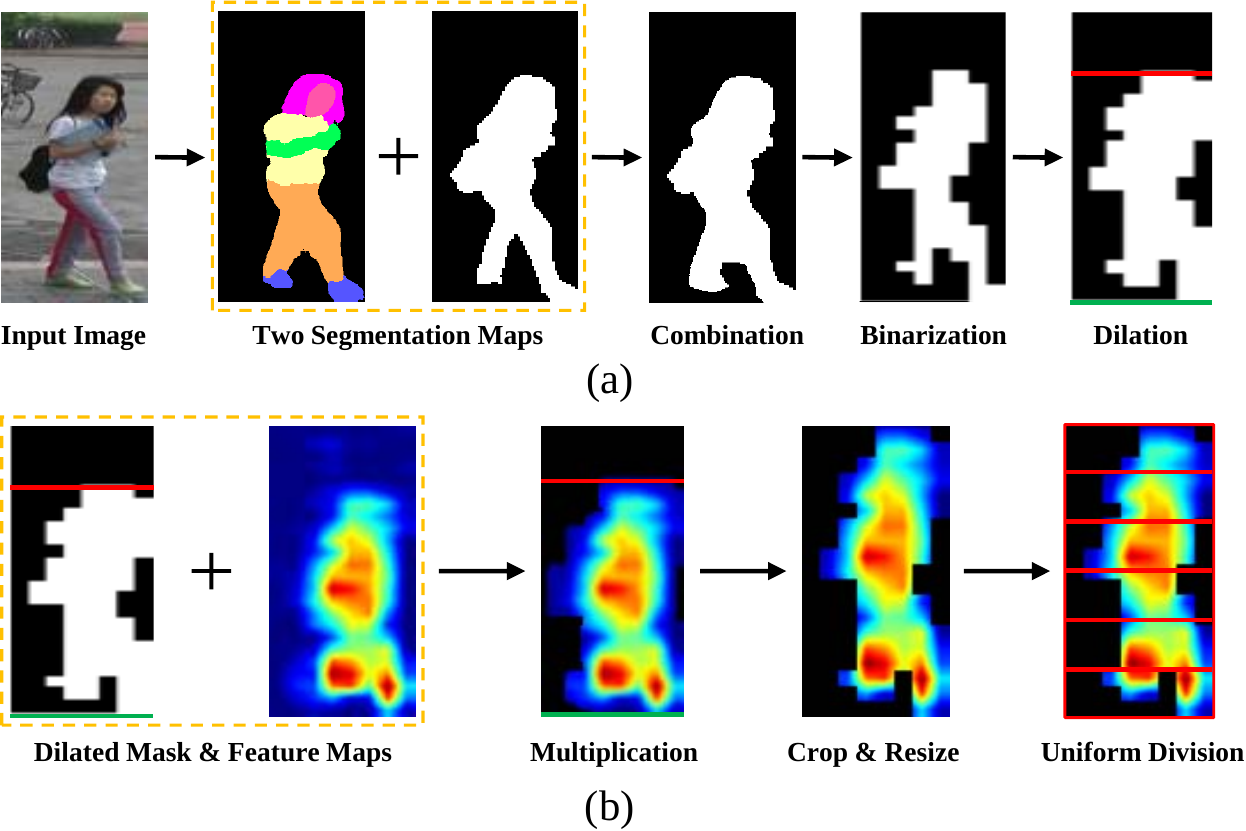}
\caption{
(a) The pipeline to obtain the coarse prior of body part locations.
The uniform division between the upper and lower boundaries of the obtained mask indicates the coarse location of each body part.
(b) The pipeline to obtain part-specific feature maps ${\bf{P}}_k (1 \leq k\leq K)$ as the input of ATs.
This process is applied to each channel of $\bf{F}$, respectively.
}
\label{fig:process}
\end{figure}

Given one training image, we propose the following pipeline to combine the outputs of both tools, as illustrated in Fig.~\ref{fig:process}(a).
First, we examine whether both the head and at least one leg are present in the parsing map by counting their respective numbers of pixels.
Second, if both are present, we obtain a more reliable mask of the human body via the union of both segmentation maps.
Third, the mask is resized to the size of the feature maps produced by the backbone model (i.e., $24\times8$ in our implementation).
We then binarize and dilate the resized mask via a $1\times2$ kernel, thus further reducing the impact of errors on human segmentation.

Similar to~\cite{12song2018mask}, the influence of background clutter is reduced through the use of the final mask $\bf{M}$.
In contrast to~\cite{12song2018mask}, we also obtain the upper and lower boundaries of the human body in the mask.
These two boundaries define the region of interest (ROI) of the human body.
The uniform division between the two boundaries in the vertical direction indicates the coarse location of $K$ body parts.
Furthermore, as shown in Fig.~\ref{fig:mask}(d-f), either the head or both legs may be absent in a small number of low-quality images;
under these circumstances, we cannot obtain the precise upper or lower boundary.
In these cases, we divide the entire image evenly in the vertical direction to estimate the coarse locations of the $K$ body parts.

\subsection{Channel Selection via Parameter Space Alignment}
\label{sec:parameterAlign}
In this subsection, we explain how MPN solves the part-relevant channel selection problem for MTs via parameter space alignment (PSA) between each MT-AT pair during training.
As illustrated in Fig.~\ref{fig:framework}, the MT and AT for each body part are built on the same backbone model, i.e., ResNet-50~\cite{16he2016deep},
and both extract part-relevant channels to construct part-level representations for ReID.
Following~\cite{5sun2017beyond}, we remove the last spatial down-sampling operation of ResNet-50 to increase the size of the output feature maps.
These feature maps are denoted as $\bf{F}$ for simplicity below.

\begin{figure}
\centering
\includegraphics[width=1.0\linewidth]{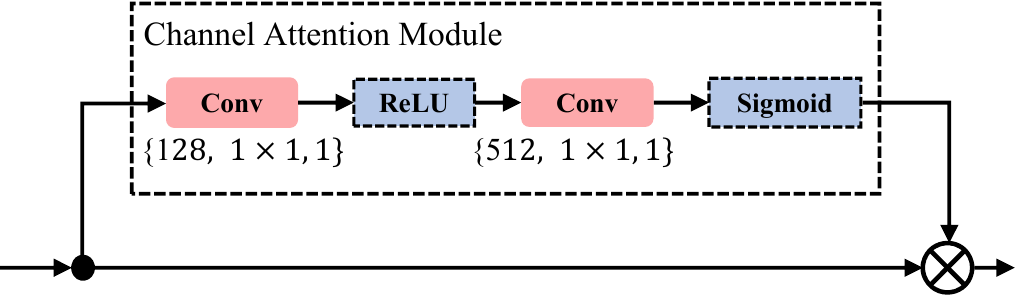}
\caption{Structure of the adopted channel attention (CA) module~\cite{15hu2018squeeze}.
The items in each bracket denote the number of filters, kernel size, and stride, respectively.
Each Conv layer is followed by a BN layer by default.}
\label{fig:ca}
\end{figure}

All MT and AT model structures are similar. Taking one MT as an example, it incorporates one $1\times1$ Conv layer, one GMP layer,
one optional channel attention (CA) module~\cite{15hu2018squeeze}, another $1\times1$ Conv layer, and one fully connected (FC) layer for classification.
Each Conv layer is followed by one batch normalization (BN)~\cite{ioffe2015batch} layer and one ReLU layer~\cite{90} by default.
The dimension of both Conv layers is set as 512.
Moreover, the configuration of the CA module is illustrated in more detail in Fig.~\ref{fig:ca}.
The loss functions for the MTs and ATs will be introduced in Sec.~\ref{loss_functions}.

The first Conv layer selects and combines part-relevant channels from $\bf{F}$.
The degree of relevance is determined by the entire training set; therefore, it may not be optimal for each individual image.
GMP transforms the feature maps into one feature vector, the elements of which are robust to the translation of body parts.
The feature vector is fed into the CA module, which overcomes the problem of the first Conv layer by recalibrating each channel according to its importance in each specific image.
The downside of using CA is that it increases the degree of model complexity; therefore, we consider this module optional in our model.
The second Conv layer projects the feature vector to a more discriminative space.

The main difference between the MT and AT of one body part has to do with their inputs.
The input of MT is the original $\bf{F}$, which means that the MT itself contains no cues for use in identifying part-relevant channels.
In comparison, we process $\bf{F}$ to obtain the part-specific feature maps ${\bf{P}}_k (1 \leq k\leq K)$,
which are the input of the $k$-th AT.
This procedure includes three steps, which are illustrated in Fig.~\ref{fig:process}(b).
First, each channel of $\bf{F}$ is multiplied by $\bf{M}$ in an element-wise manner, reducing the impact of background clutter.
Second, the portion of feature maps within the upper and lower boundaries are resized to the original size of $\bf{F}$ via bilinear interpolation;
this step corrects errors in pedestrian detection.
Finally, the uniform division of the resized feature maps in the vertical direction produces $K$ part-specific feature maps.

As illustrated in Fig.~\ref{fig:channelResponse}, channels in $\bf{F}$ activate local responses at their corresponding body parts.
This means that only relevant channels have high responses in ${\bf{P}}_k$ following the division operation;
therefore, it is much easier to optimize the first Conv layer of AT than it is to optimize MT.
The parameters of this layer can create inductive bias, assisting MT in selecting part-relevant channels.
We propose to utilize this inductive bias by simply sharing the parameters of the first $1\times1$ Conv layer between each MT-AT pair.

However, sharing parameters for channel selection alone cannot ensure that only the part-relevant channels will be selected for MT,
because MT itself contains no cues for part-relevant channel selection.
When MT and AT optimize the shared Conv layer together, irrelevant channels may also be selected.
This means that there is a gap between the features extracted by MT and those extracted by AT.
To resolve this problem, we apply stronger regularization by further sharing the parameters of the CA module and
the second $1\times1$ Conv layer between MT and AT, respectively.
By sharing parameters of each respective layer, we implicitly require that its input feature vectors from MT and AT will be similar for each image.
This constraint, in turn, regularizes the first $1\times1$ Conv layer to select part-relevant channels.

In conclusion, MPN shares the parameters of the two Conv layers and the CA module between each MT-AT pair, respectively.
We refer to this hard parameter sharing strategy as PSA,
which forces the first $1\times1$ Conv layer to select part-relevant channels for ReID purposes.
In other words, MT is regularized to extract semantically aligned part-level representations.
Compared to existing works~\cite{8zhang2018densely,ren2018factorized},
one important advantage of hard parameter sharing is that it makes the MPN architecture very compact in the training stage.
During testing, all ATs are removed, meaning that MPN is free from body part detection after training is complete.

\subsection{Respective Loss Functions for MTs and ATs}
\label{loss_functions}
We employ two popular loss functions to train MTs and ATs.
First, we attach one cross-entropy loss function to the classification layer of each MT and AT, respectively:
\begin{equation}
\mathcal{L}_{ID} = - \frac{1}{N}\sum_{l = 1}^{N}\sum_{k = 1}^{K}
\left(\mathcal{L}_{ce}\left({\bf{W}}_{k}^{mt}{\bf{f}}_{k}^{l}\right)+\mathcal{L}_{ce}\left({\bf{W}}_{k}^{at}{\bf{z}}_{k}^{l}\right)\right),
\end{equation}
where $N$ denotes batch size.
${\bf{f}}_{k}^{l}$ and ${\bf{z}}_{k}^{l}$ represent the feature vectors extracted from the $k$-th part of the $l$-th image by MTs and ATs, respectively.
${\bf{W}}_{k}^{mt}$ and ${\bf{W}}_{k}^{at}$ represent parameters of the classification layers for the $k$-th MT and AT, respectively.
As illustrated in Fig.~\ref{fig:framework}, ${\bf{f}}_{k}^{l}$ and ${\bf{z}}_{k}^{l}$ are the outputs of the second Conv layer of MTs and ATs, respectively.
$\mathcal{L}_{ce}$ stands for the cross-entropy loss function.

The $K$ part-level features extracted by MTs are concatenated as the holistic representation $\bf{h}$ of one image:
\begin{equation}
{\bf{h}} = [{{\bf{f}}_{1}},{{\bf{f}}_{2}}, ..., {{\bf{f}}_{K}}].
\label{Eq:h}
\end{equation}

We then apply the triplet loss~\cite{75hermans2017defense} to ensure that the distance between the representations
of intra-class image pairs is smaller than that of the inter-class image pairs.
To ensure that sufficient triplets are sampled, we randomly choose $A$ images in each of $S$ random subjects to create a mini-batch.
We follow the~\emph{BatchHard} strategy used in~\cite{75hermans2017defense} to sample the triplets.
The triplet loss can be formulated as $\mathcal{L}_{TRI}=$
\begin{equation}
\frac{1}{N_T} \sum_{i = 1}^{S} \sum_{a = 1}^{A}[ \max \limits_{\substack{p=1...A}} \mathcal{D}\left( {\bf{h}}^{a}_{i}, {\bf{h}}^{p}_{i} \right) - \min \limits_{\substack{{n=1...A} \\ {j=1...S} \\ j \neq i }} \mathcal{D}\left({\bf{h}}^{a}_{i},{\bf{h}}^{n}_{j} \right) + \alpha]_{+},
\label{eqtrir}
\end{equation}
where $\{{\bf{h}}^{a}_{i}, {\bf{h}}^{p}_{i}, {\bf{h}}^{n}_{j}\}$ compose a triplet.
The anchor and positive images are sampled from the $i$-th subject, while the negative image is sampled from the $j$-th subject.
$\alpha$ denotes the margin of the triplet constraint and $N_T$ represents the number of triplets in a batch that violate the triplet constraint~\cite{75hermans2017defense}. $\left[*\right]_{+} = max(0, *)$ is the hinge loss.
$\mathcal{D}\left( {\bf{h}}^{a}_{i}, {\bf{h}}^{p}_{i} \right)$ and $\mathcal{D}\left({\bf{h}}^{a}_{i},{\bf{h}}^{n}_{j} \right)$ denote the cosine distance between two feature vectors. For example,
\begin{equation}
\mathcal{D}\left( {\bf{h}}^{a}_{i}, {\bf{h}}^{p}_{i} \right)=1-\frac{{{\bf{h}}^{a}_{i}}^{T}{\bf{h}}^{p}_{i}}{\|{\bf{h}}^{a}_{i}\|  \|{\bf{h}}^{p}_{i}\|}.
\label{eq:cosDistance}
\end{equation}

Note that the importance of ATs and MTs are asymmetric. ATs are removed during testing; therefore, we do not apply the triplet loss to the features extracted by ATs.

\subsection{Feature Space Alignment between MTs and ATs}
The above MTL strategy enables MPN to achieve strong performance.
However, due to the difference between the input feature maps of each MT-AT pair, a gap still exists between the features extracted by MTs and ATs.
Accordingly, to bridge this discrepancy, we propose the following method to align their features in a class-wise manner.

First, the $K$ part-level features extracted by ATs are concatenated to form another holistic representation ${\bf{g}}$ of one image:
\begin{equation}
{\bf{g}} = [{{\bf{z}}_{1}},{{\bf{z}}_{2}}, ..., {{\bf{z}}_{K}}].
\end{equation}

Second, we calculate the mean representations for each subject in one batch:
\begin{equation}
\overline{\bf{h}}_{i} = \frac{1}{A} \sum_{a = 1}^{A} {\bf{h}}^{a}_{i},
\label{eq:meanH}
\end{equation}
and,
\begin{equation}
\overline{\bf{g}}_{i} = \frac{1}{A} \sum_{a = 1}^{A} {\bf{g}}^{a}_{i}.
\label{eq:meanG}
\end{equation}

Finally, we penalize the cosine distance between $\overline{\bf{h}}_{i}$ and $\overline{\bf{g}}_{i}$:
\begin{equation}
\mathcal{L}_{CF}=\frac{1}{S} \sum_{i = 1}^{S} \mathcal{D}\left(\overline{\bf{h}}_{i}, \overline{\bf{g}}_{i} \right).
\label{eq:MeanDist}
\end{equation}

Our class-wise feature alignment strategy can be contrasted with the sample-wise feature alignment approaches in recent works~\cite{ren2018factorized,8zhang2018densely}.
For example, consider the approach adopted in~\cite{ren2018factorized}:
\begin{equation}
\mathcal{L}_{SF}=\frac{1}{N} \sum_{i = 1}^{S} \sum_{a = 1}^{A} \mathcal{D}\left({\bf{h}}^{a}_{i}, {\bf{g}}^{a}_{i} \right).
\label{eq:SampleDist}
\end{equation}

Compared with $\mathcal{L}_{CF}$, $\mathcal{L}_{SF}$ imposes a stronger constraint,
as it requires that the distance between each pair of ${{\bf{h}}^{a}_{i}}$ and ${\bf{g}}^{a}_{i}$ should be minimized.
This requirement is reasonable when the quality of ${\bf{g}}^{a}_{i}$ is very high;
in practice, however, the quality of ${\bf{g}}^{a}_{i}$ is limited due to errors in the body part location priors.
First, the outside tools adopted in this paper and in~\cite{8zhang2018densely} may fail for low-quality pedestrian images.
Second, the prior that we employ in Sec.~\ref{sec:part_prior} is coarse.
The uniform division operation on the human body presented in Fig.~\ref{fig:process}(b) may not account for dramatic movement of flexible body parts in one image.

Therefore, $\mathcal{L}_{CF}$ is a more robust constraint. By penalizing the distance between the class-wise mean representations of MTs and ATs,
$\mathcal{L}_{CF}$ becomes less vulnerable to errors in body part location priors.
In the experimental section, we will demonstrate that $\mathcal{L}_{CF}$ outperforms both constraints proposed in~\cite{ren2018factorized,8zhang2018densely}.

\section{Person ReID via MPN}
In the training stage, the overall objective function of MPN can be formulated as follows:
\begin{equation}
\begin{aligned}
\mathcal{L} = \mathcal{L}_{ID}+\mathcal{L}_{TRI}+\lambda \mathcal{L}_{CF},
\end{aligned}
\end{equation}
where $\lambda$ is a weight term.

In the testing stage, all ATs are removed. We employ ${\bf{h}}$ in Eq.~\ref{Eq:h} as the representation of one image.
We consistently adopt the cosine metric to measure the similarity $\rho$ between two representations ${\bf{h}}_{1}$ and ${\bf{h}}_{2}$:
\begin{equation}
\rho = \frac{{\bf{h}}_{1}^{T}{\bf{h}}_{2}}{\|{\bf{h}}_{1}\|  \|{\bf{h}}_{2}\|}.
\end{equation}

\section{Experiments}
In this section, we conduct comprehensive experiments on four publicly available large-scale benchmark datasets: Market-1501 \cite{27zheng2015scalable},
DukeMTMC-ReID \cite{59zheng2017unlabeled}, CUHK03 \cite{33li2014deepreid}, and MSMT17 \cite{80PTGAN}.
We follow the official evaluation protocols for each of these databases and further adopt both Rank-1 accuracy and mean Average Precision (mAP) as evaluation metrics for all benchmarks.

The Market-1501 database~\cite{27zheng2015scalable} consists of 32,668 pedestrian images captured by six cameras of 1,501 identities.
The Deformable Part Model (DPM)~\cite{felzenszwalb2008discriminatively} is employed to detect bounding boxes for these pedestrians.
Market-1501 is divided into a training set and a testing set:
the former includes 12,936 images of 751 identities, while the latter comprises images of the remaining 750 identities.
Moreover, the testing set is further split into a gallery set and a query set, which contain 19,732 and 3,368 images, respectively.

The DukeMTMC-ReID database~\cite{59zheng2017unlabeled} contains 36,441 pedestrian images of 1,404 identities.
The images were captured by eight high-resolution cameras.
A total of 16,522 images of 702 identities make up the training set, while images of the other 702 identities make up the testing set.
The testing set is further split into a gallery set, containing 17,661 images, and a query set, containing the remaining 2,268 images.

The CUHK03 database~\cite{33li2014deepreid} consists of 14,097 pedestrian images of 1,467 identities.
The images were captured by two disjoint cameras.
The bounding boxes of pedestrians in CUHK03 are obtained by means of two methods, namely human annotation and DPM detection.
We report results using each of these two types of bounding boxes.
We adopt the training/testing splitting protocol proposed in~\cite{68zhong2017re}.
In this protocol, images of 767 identities are used for training, while images of the remaining 700 identities are utilized for testing.

The MSMT17 database~\cite{80PTGAN} contains 126,441 pedestrian images of 4,101 identities in total.
This dataset was collected by a camera network comprising 12 outdoor cameras and three indoor ones.
Faster R-CNN~\cite{81faster} is used for pedestrian detection.
MSMT17 is split into a training set, containing 32,621 images of 1,041 identities,
and a testing set, consisting of 93,820 images of 3,060 identities.
Furthermore, the testing set is randomly divided into a gallery set and a query set, which consist of 82,161 and 11,659 images respectively.
Compared with the above datasets, MSMT17 is more challenging, because its scale is larger and it includes more complex background and illumination changes.

\subsection{Implementation Details}
Firstly, all images in the above four databases are resized to $384\times128$ pixels. Example images can be found in Fig.~\ref{fig:partMisalignment} and Fig.~\ref{fig:channelResponse}. Data augmentation is utilized to reduce overfitting in the training stage.
First, offline translation~\cite{50zhao2014learning} is adopted to enlarge each training set by a factor of five.
Second, random erasing~\cite{54zhong2017random} and horizontal flipping with a ratio of 0.5 are utilized for online augmentation.
We set $S$ as 6 and $A$ as 8 to construct a mini-batch; thus, the batch size is 48.

\begin{figure}
\centering
\includegraphics[width=1.0\linewidth]{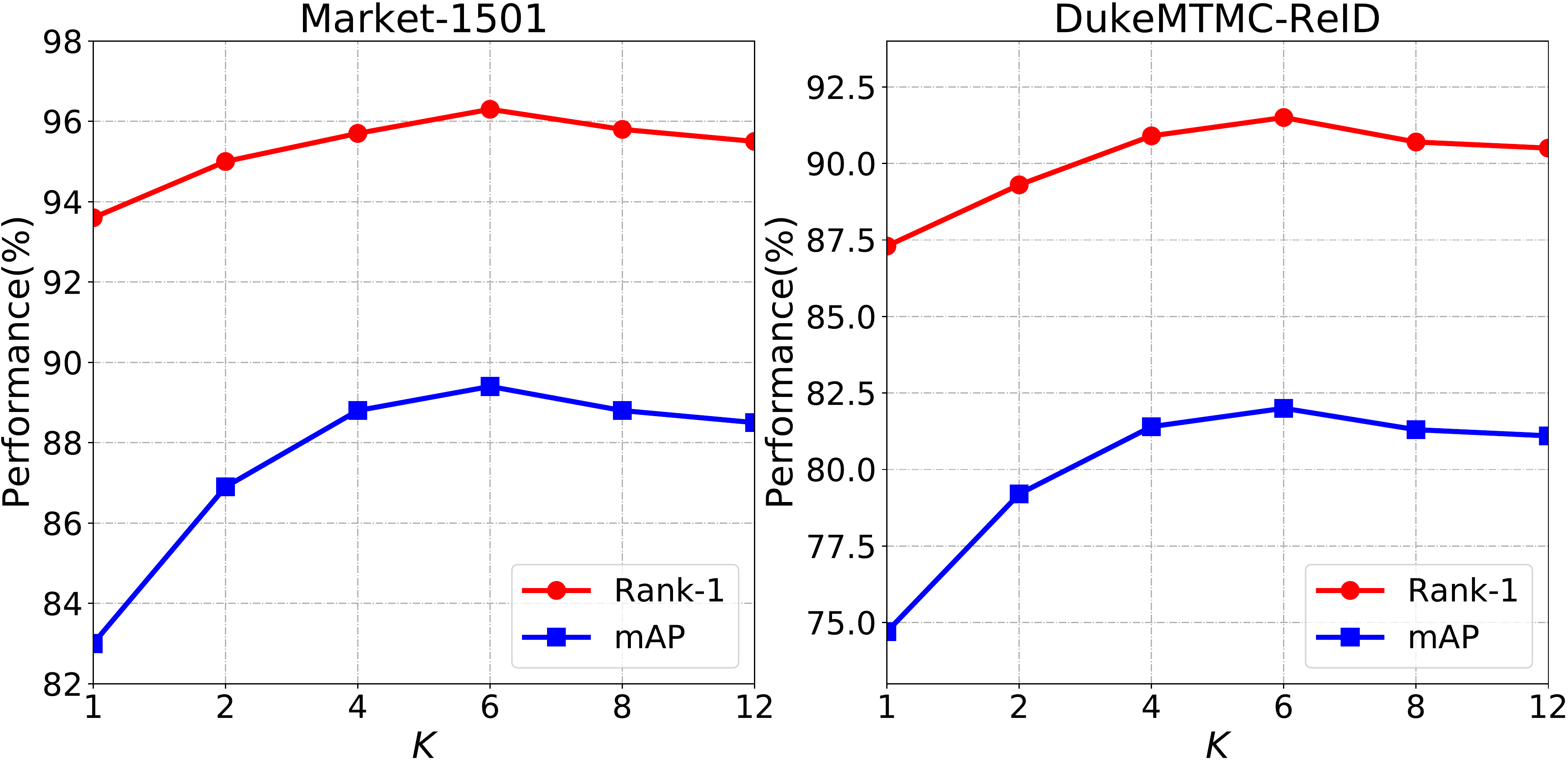}
\caption{Evaluation on the value of hyper-parameter $K$ for the performance of MPN.
}
\label{fig:PartNum}
\end{figure}

There are only a few hyper-parameters for MPN.
We empirically set $K$ as 6, according to the evaluation results in Fig~\ref{fig:PartNum}.
$\alpha$ and $\lambda$ are consistently set to 0.2 and 1 respectively for the sake of simplicity.
The PyTorch framework is used for implementation.
The standard stochastic gradient descent (SGD) optimizer, with a weight decay of $5 \times 10^{-4}$, is utilized for model optimization.
The momentum \cite{31sutskever2013importance} value is set as 0.9.
The parameters of MPN are initialized from those of the IDE model~\cite{28zheng2017person} trained on each respective database;
subsequently, MPN is trained in an end-to-end fashion for $70$ epochs.
The learning rate is initially set to 0.01, then multiplied by 0.1 for every 20 epochs.

\subsection{Ablation Study}
In the following, we systematically investigate the effectiveness of each key component of MPN:
namely, MTL structure, along with the parameter and feature space alignment between each MT-AT pair, respectively.
Experiments are conducted on three popular databases: Market-1501, DukeMTMC-ReID, and CUHK03.
Results are summarized in Table~\ref{tab:ablation}.

\newcommand{\tabincell}[2]{\begin{tabular}{@{}#1@{}}#2\end{tabular}}
\begin{table*}
\caption{Ablation Study on Each Component of MPN}
\centering
\begin{center}
\begin{tabular}{p{2.5cm}<{\centering}|p{0.45cm}<{\centering} p{0.6cm}<{\centering} p{0.6cm}<{\centering} p{0.3cm}<{\centering} p{0.65cm}<{\centering}|cc|cc|cc|cc}
\hline
  Dataset & \multicolumn{5}{c|}{Components} & \multicolumn{2}{c|}{Market-1501} & \multicolumn{2}{c|}{DukeMTMC-ReID} &\multicolumn{2}{c|}{CUHK03-detected} &\multicolumn{2}{c}{CUHK03-labeled} \\
  \hline
  Metric        &MTL & C1-S & C2-S &CF   &CA  &Rank-1& mAP  &Rank-1& mAP    &Rank-1&mAP   &Rank-1&mAP \\
  \hline
  \hline
  Baseline      &-   &-     &-     &-    &-    &94.2&84.4   &88.2&77.4    &70.9&66.7    &75.6&71.3\\
  Baseline (UB) &-   &-     &-     &-    &-    &94.9&85.8   &88.9&78.9    &75.6&71.3    &-&-\\
  MT Only       &-   &-     &-     &-    &-    &94.4&85.6   &88.6&78.0    &74.1&69.3    &78.7&73.9\\
  Na\"ive MTL     &\checkmark  &-  &-  &-  &-    &94.6&87.2   &88.7&78.8    &75.5&70.7    &78.9&75.1\\
  \hline
  \multirow{3}*{\tabincell{c}{PSA Alone}}
                                 &\checkmark  &\checkmark &-          &-          &-          &95.5&88.5   &90.4&80.7   &81.1&76.9   &83.0&79.1\\
  ~                              &\checkmark  &-          &\checkmark &-          &-          &95.1&88.2   &90.1&80.1   &80.6&76.2   &82.5&78.5\\
  ~                              &\checkmark  &\checkmark &\checkmark &-          &-          &95.8&88.7   &90.8&80.9   &81.6&77.1   &83.4&79.5\\
  \hline
  \multirow{2}*{\tabincell{c}{FSA Alone}}
                                 &\multirow{2}*{ \checkmark }
                                 &\multirow{2}*{-}
                                 &\multirow{2}*{-}
                                 &\multirow{2}*{ \checkmark }
                                 &\multirow{2}*{-}
                                 &\multirow{2}*{95.5}&\multirow{2}*{88.3}
                                 &\multirow{2}*{90.5}&\multirow{2}*{80.7}
                                 &\multirow{2}*{81.0}&\multirow{2}*{76.9}
                                 &\multirow{2}*{82.6}&\multirow{2}*{79.1}   \\
  ~ & & & & & & & & & & & & & \\
  \hline
  MPN$^o$                        &\checkmark  &\checkmark &\checkmark &\checkmark &-           &96.1&89.2   &91.2&81.6   &82.6&78.4   &84.1&80.3\\
  MPN                            &\checkmark  &\checkmark &\checkmark &\checkmark &\checkmark  &96.3&89.4   &91.5&82.0   &83.4&79.1   &85.0&81.1\\
  \hline
\end{tabular}
\end{center}
\label{tab:ablation}
\end{table*}

\subsubsection{Effectiveness of Na\"ive Multi-task Learning}
We first evaluate the performance of one na\"ive MTL approach. In this approach, both PSA and FSA are removed from MPN.
To facilitate a clean comparison, the models in this experiment are not equipped with the CA module for either MTs or ATs.
The other details of the model architecture and training strategy remain the same as those in MPN.
We compare its performance with two basic methods.

The first method is similar to the popular Part-based Convolutional Baseline (PCB) approach~\cite{5sun2017beyond}.
In this method, only ATs are reserved in MPN for both the training and the testing stages.
Following~\cite{5sun2017beyond}, $\bf{F}$ is uniformly divided into $K$ horizontal stripes, which are used as the input of ATs.
Accordingly, this method is incapable of handling the body part misalignment problem.
Moreover, to facilitate fair comparison with the na\"ive MTL approach,
we also concatenate the $K$ part-level features produced by ATs and add triplet loss in the training stage.
In the testing stage, the concatenated part-level features are used as the representation of one image.
This method is employed as the baseline in this paper.
In the second method, only MTs are reserved in both training and testing stages;
thus, there is no guidance for MTs to learn part-specific representations.
The other details remain the same as MPN. This method is denoted as MT Only in Table~\ref{tab:ablation}.

We also show the upper bound of the baseline's performance, which is referred to as Baseline (UB) in Table~\ref{tab:ablation}.
In Baseline (UB), we first correct pedestrian detection errors for both training and testing images,
and then test the performance of the baseline.
For the first two databases, the detection errors are corrected according to the scheme in Fig~\ref{fig:process}.
For CUHK03, we can directly report the baseline's performance on the CUHK03-labeled dataset, where pedestrian detection was manually performed.

From the comparison results in Table~\ref{tab:ablation}, it can be seen that MT Only outperforms the baseline approach.
This may be because MT Only is not affected by the body part misalignment problem in the spatial dimension.
In comparison, the uniform division operation in baseline is sensitive to the subtle change of body part locations.
Moreover, the na\"ive MTL approach consistently outperforms both basic methods;
this is because ATs can regularize the backbone model in order to learn more diverse local features in $\bf{F}$~\cite{6yao2017deep},
which both relieves the overfitting problem and enables MTs to extract stronger representations.
The above results verify the effectiveness of the na\"ive MTL structure.

\subsubsection{Effectiveness of PSA}
In this experiment, FSA is removed from MPN, and the effectiveness of hard parameter sharing for PSA is evaluated. The CA modules are removed to facilitate clean comparison (in a similar way to the above experiment).
In Table~\ref{tab:ablation}, C1-S and C2-S denote whether the parameters of the first and the second $1\times1$ Conv layer are shared between each MT-AT pair, respectively.

Experimental results in Table~\ref{tab:ablation} demonstrate that both C1-S and C2-S can further improve the performance of na\"ive MTL by a considerable margin.
For example, C1-S outperforms the na\"ive MTL in terms of Rank-1 accuracy by 0.9\%, 1.7\%, 5.6\%, and 4.1\% on each database in Table~\ref{tab:ablation}, respectively.
Moreover, performance promotion via C1-S is more significant than that achieved by C2-S.
The above results verify the importance of sharing the first $1\times1$ Conv layer between each MT-AT pair for part-relevant channel selection.

Finally, by sharing both $1\times1$ Conv layers, stronger performance is consistently achieved on all databases;
this indicates that C1-S and C2-S are complementary. Compared with the baseline approach,
MTL with PSA achieves significantly better performance.
In particular, the mAP is promoted by 4.3\%, 3.5\%, 10.4\%, and 8.2\% respectively on each database in Table~\ref{tab:ablation}.
The above experimental results justify the effectiveness of PSA for part-aware ReID.

\subsubsection{Effectiveness of FSA}
In this experiment, PSA is removed from MPN so that we can investigate the effectiveness of the proposed class-wise feature alignment strategy (abbreviated as CF in Table~\ref{tab:ablation}).
All CA modules are removed from MPN to facilitate clean comparison.
As shown in Table~\ref{tab:ablation}, MTL with FSA consistently outperforms the baseline approach by a large margin:
in brief, Rank-1 accuracy is improved by 1.3\%, 2.3\%, 10.1\%, and 7.0\%,
while mAP is also promoted by 3.9\%, 3.3\%, 10.2\%, and 7.8\% on each benchmark, respectively.
These experimental results verify the effectiveness of the proposed FSA method.

\subsubsection{Combination of PSA and FSA}
The next step is to combine the parameter and feature space alignments together.
Again, to ensure fair comparison with the above results, we test the performance of MPN without the CA modules (denoted as MPN$^o$ in Table~\ref{tab:ablation}).
As shown in Table~\ref{tab:ablation}, MPN$^o$ consistently outperforms all models in the above experiments;
this indicates that parameter and feature space alignments work in a complementary fashion to help MTs learn semantically aligned part-level features.
The performance achieved by MPN$^o$ is significantly higher than that obtained using the baseline approach.
For example, MPN$^o$ outperforms the baseline approach by 4.8\%, 4.2\%, 11.7\%, and 9.0\%, respectively, in terms of mAP on each dataset in Table~\ref{tab:ablation}.
In particular, pedestrian detection was manually performed on the CUHK03-labeled database, which means that there are rarely pedestrian detection errors.
Comparisons on this database indicate that MPN effectively handles the body part misalignment problem caused by other factors, e.g. pose variation.
Besides, it is worth noting that MPN$^o$ is both powerful and compact.
Compared with the baseline model, moreover, MPN$^o$ has more parameters only on the extra classification layers of ATs in the training stage.
In the testing stage, the number of parameters for MPN$^o$ and the baseline is exactly the same.

Finally, we equip MPN$^o$ with CA modules; this is denoted as MPN in Table~\ref{tab:ablation}.
In our implementation, we share the parameters of the CA module for each MT-AT pair, respectively.
Experimental results indicate that implementation of the CA modules results in the consistently improved performance of MPN on all four databases.
This is because the role of the CA modules is complementary to that of the first $1\times1$ Conv layers in MTs, as explained in Sec.~\ref{sec:parameterAlign}.

\subsubsection{Visualization of Attention Maps for MTs}
\begin{figure}
\centering
\includegraphics[width=1.0\linewidth]{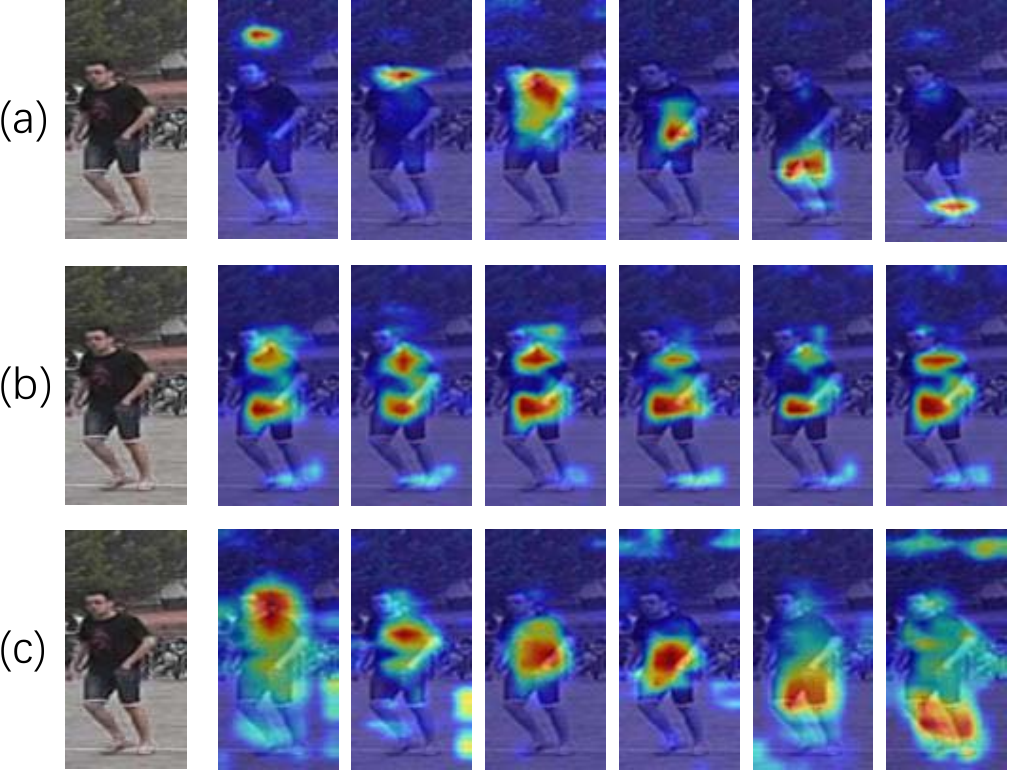}
\caption{Visualization of attention maps for each of the $K$ classifiers' prediction using Grad-CAM. Three representative models in Table~\ref{tab:ablation} are compared:
(a) baseline model; (b) na\"ive MTL; (c) MPN. (Best viewed in color.)
}
\label{fig:threeModel_grad}
\end{figure}

We further support the above experimental results by visualizing the attention maps for each of the $K$ classifiers' prediction using Grad-CAM~\cite{77grad}.
Three representative models in Table~\ref{tab:ablation} are compared: the baseline, na\"ive MTL, and MPN.
The attention maps on one pedestrian image with a misalignment problem are illustrated in Fig.~\ref{fig:threeModel_grad}.

The following observations can be made.
First, attention maps for the baseline model are not reasonable for the image in Fig.~\ref{fig:threeModel_grad}.
For example, attention for the first classifier focuses on the background area.
This is because the uniform division operation on the feature maps brings about the semantic misalignment problem for the $K$ classifiers.
Second, in the absence of any guidance, the $K$ attention maps for the na\"ive MTL approach are very similar,
which means it lacks diversity for the features extracted by the $K$ MTs in the na\"ive MTL model.
Third, with the guidance of the proposed dual alignment strategies, MPN can learn reasonable spatial localization of body parts;
therefore, part-level features extracted by MPN are well aligned in semantics.

\begin{figure}
\centering
\includegraphics[width=0.95\linewidth]{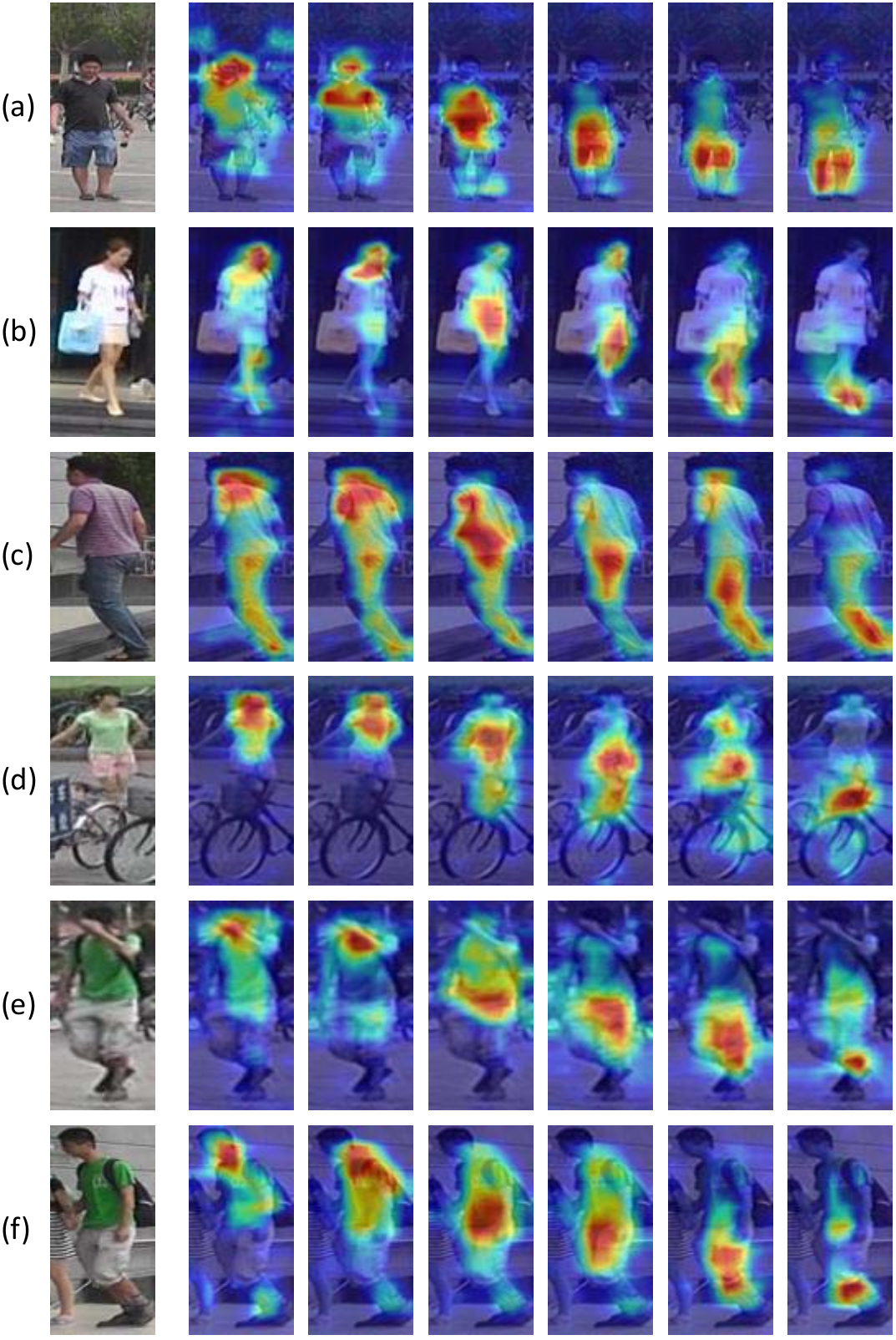}
\caption{Visualization of attention maps for each MT's classifier of MPN using Grad-CAM.
The attention maps of MPN are semantically consistent across images in the face of various challenges,
e.g. errors in pedestrian detection (a, b), dramatic pose variations (c, e), background clutter and occlusion (d).
(Best viewed in color.)
}
\label{fig:mpn_grad}
\end{figure}

Furthermore, in Fig.~\ref{fig:mpn_grad}, we visualize the attention maps for each MT's classifier of MPN on more images in the Market-1501 database.
It can thereby be seen that the attention maps are both reasonable and semantically consistent when faced with various types of challenges,
e.g. errors in pedestrian detection (Fig.~\ref{fig:mpn_grad}a and Fig.~\ref{fig:mpn_grad}b),
dramatic pose variations (Fig.~\ref{fig:mpn_grad}c and Fig.~\ref{fig:mpn_grad}e), background clutter and occlusion (Fig.~\ref{fig:mpn_grad}d).
In particular, MPN can extract features robustly from the upper arms and legs, regardless of pose variations.
However, the lower arms are ignored in Fig.~\ref{fig:mpn_grad}. This is because pedestrians in Market-1501 wear short sleeves.
There are no clothes on the lower arms and therefore the lower arms lack discriminative power.
In the supplementary file, we show MPN can accurately attend to the lower arms if pedestrians in one database also wear long sleeves.

The ability of MPN that attends to flexible body parts can be explained as follows.
Let us take the upper arms as an example, they are in a similar pose in most images.
According to our definition of body parts in Fig.~\ref{fig:process}(b), most upper arms lie in the second body part region,
which means we obtain good priors of upper-arm locations for most images.
Therefore, the first Conv layer in the second MT-AT pair will select channels corresponding to upper arms for part-level feature extraction.
The same explanation applies to the lower arms and legs.
The above analysis proves that MPN can extract semantically aligned part-level features in a robust manner.

\subsection{Comparisons with Variants of MPN}
We now compare the performance of MPN with some of its possible variants.
In the interests of efficient evaluation, experiments are conducted on the Market-1501 and DukeMTMC-ReID databases only.
Moreover, models in this subsection are not equipped with the CA modules to facilitate clean comparison.

\subsubsection{Comparisons with Variants of the Prior Information}
We compare the performance of the proposed priors of body part locations in Sec.~\ref{sec:part_prior} with two possible variants, namely Uniform Division and ROI Resize.
The other implementation details of MPN are kept the same for the different priors.
Uniform Division means that $\bf{F}$ of each training image is uniformly divided into $K$ horizontal stripes, which form the input of ATs.
For ROI Resize, the element-wise multiplication step in Fig.~\ref{fig:process}(b) is skipped,
while the other operations remain the same as those in Fig.~\ref{fig:process}.
We also compare with the MT Only approach in Table~\ref{tab:ablation}, which adopts no prior information of body part locations.

Results of the comparison are presented in Table~\ref{tab:priorCompare}.
From the table, it can be seen that with any prior in Table~\ref{tab:priorCompare}, MPN significantly outperforms the models that adopt no prior information.
Interestingly, the coarse prior of Uniform Division also brings about a noticeable improvement.
This occurs for two main reasons. First, the pedestrian detection error for most training images is slight or moderate;
as a result, Uniform Division produces good priors for the well-aligned training images.
Second, the two alignment strategies in MPN are robust to errors in the priors.
The above analysis indicates that MPN can work well with coarse priors, which are easy to obtain in practice.

We can also observe that the more accurate the prior, the better the performance.
For example, ROI Resize rescales the human body to a canonical size and position.
As it thereby corrects the majority of misalignment errors, it achieves better performance than Uniform Division.
Moreover, the element-wise multiplication operation in Fig.~\ref{fig:process} suppresses the background clutter around body parts,
meaning that it is also helpful in promoting the performance of MPN.

\begin{table}
\caption{Performance Comparison of Different Types of Prior for Body Part Locations in the Training Stage}
\centering
\begin{tabular}{c|cc|cc}
\hline
  Dataset           & \multicolumn{2}{c|}{Market-1501} & \multicolumn{2}{c}{DukeMTMC-ReID} \\
  \hline
  Metric            & Rank-1  & mAP   & Rank-1  & mAP \\
  \hline
  \hline
  Baseline          &94.2     &84.4   &88.2     &77.4 \\
  MT Only           &94.4     &85.6   &88.6     &78.0 \\
  \hline
  Uniform Division  &95.6     &88.8   &90.8     &81.2 \\
  ROI Resize        &95.8     &89.0   &91.0     &81.3 \\
  \hline
  ours              &96.1     &89.2   &91.2     &81.6 \\
  \hline
\end{tabular}
\label{tab:priorCompare}
\end{table}

\subsubsection{Comparisons with Variant for PSA}
We constrain the parameter space of MTs via hard parameter sharing with ATs. This strategy results in a compact and efficient model in the training stage.
One natural alternative is soft parameter sharing~\cite{ruder2017overview}, which constrains the parameters between each MT-AT pair to be similar rather than identical.
To facilitate clean comparison, we compare their performance on one of the two $1\times1$ Conv layers each time and do not apply any constraints to the other layer.
Constraints in the feature space are also removed.
For soft parameter sharing, we utilize the $L2$ loss to penalize the distance between the parameters of each MT-AT pair, respectively.
Four representative values (i.e., 0.01, 0.1, 1, and 10) are used as the weights for the $L2$ loss, respectively.
By contrast, hard parameter sharing does not include hyper-parameters.

\begin{figure}
\centering
\subfigure[]{
\begin{minipage}[b]{0.47\textwidth}
\includegraphics[width=1\textwidth]{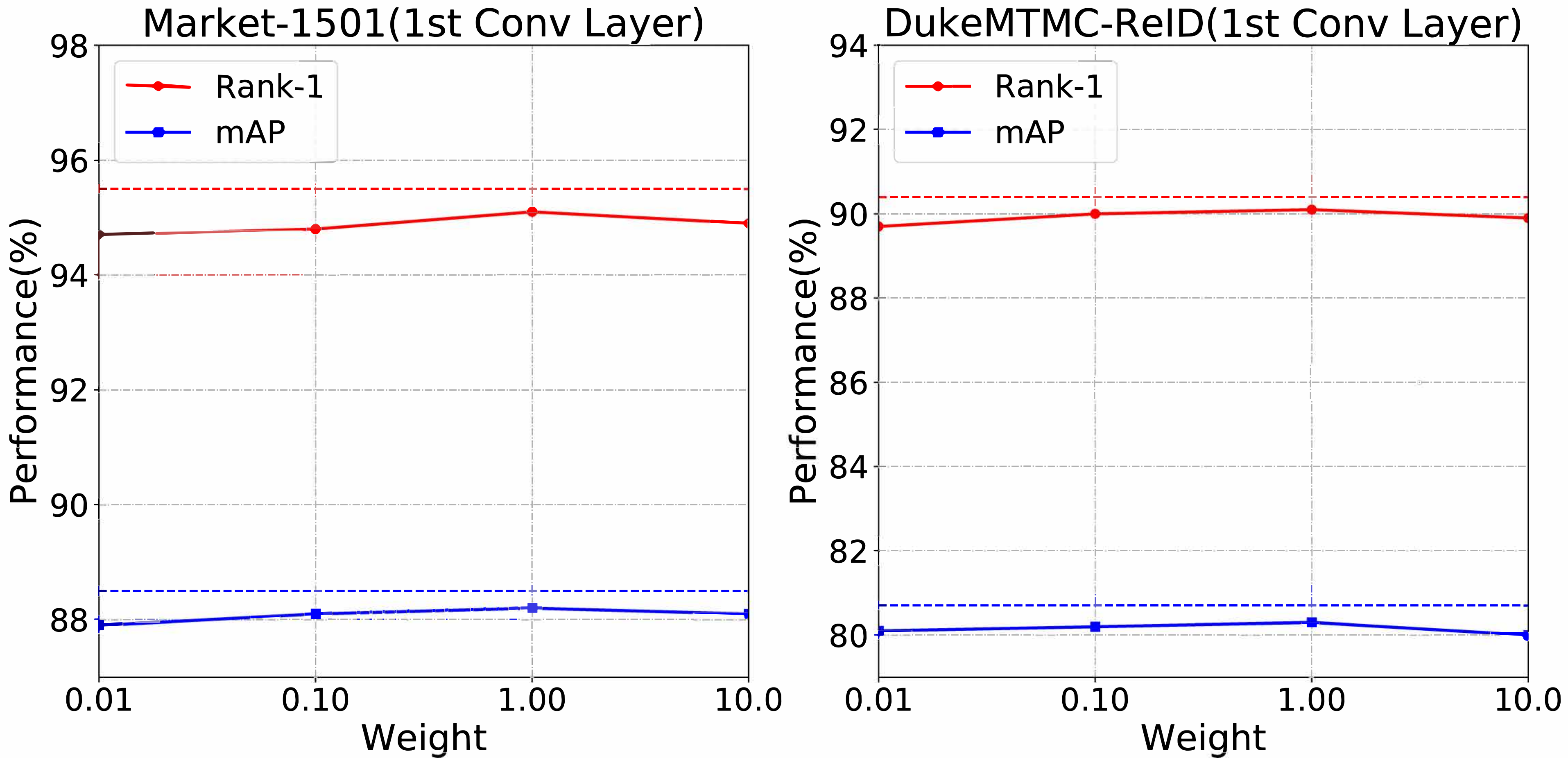}
\end{minipage}
}
\subfigure[]{
\begin{minipage}[b]{0.47\textwidth}
\includegraphics[width=1\textwidth]{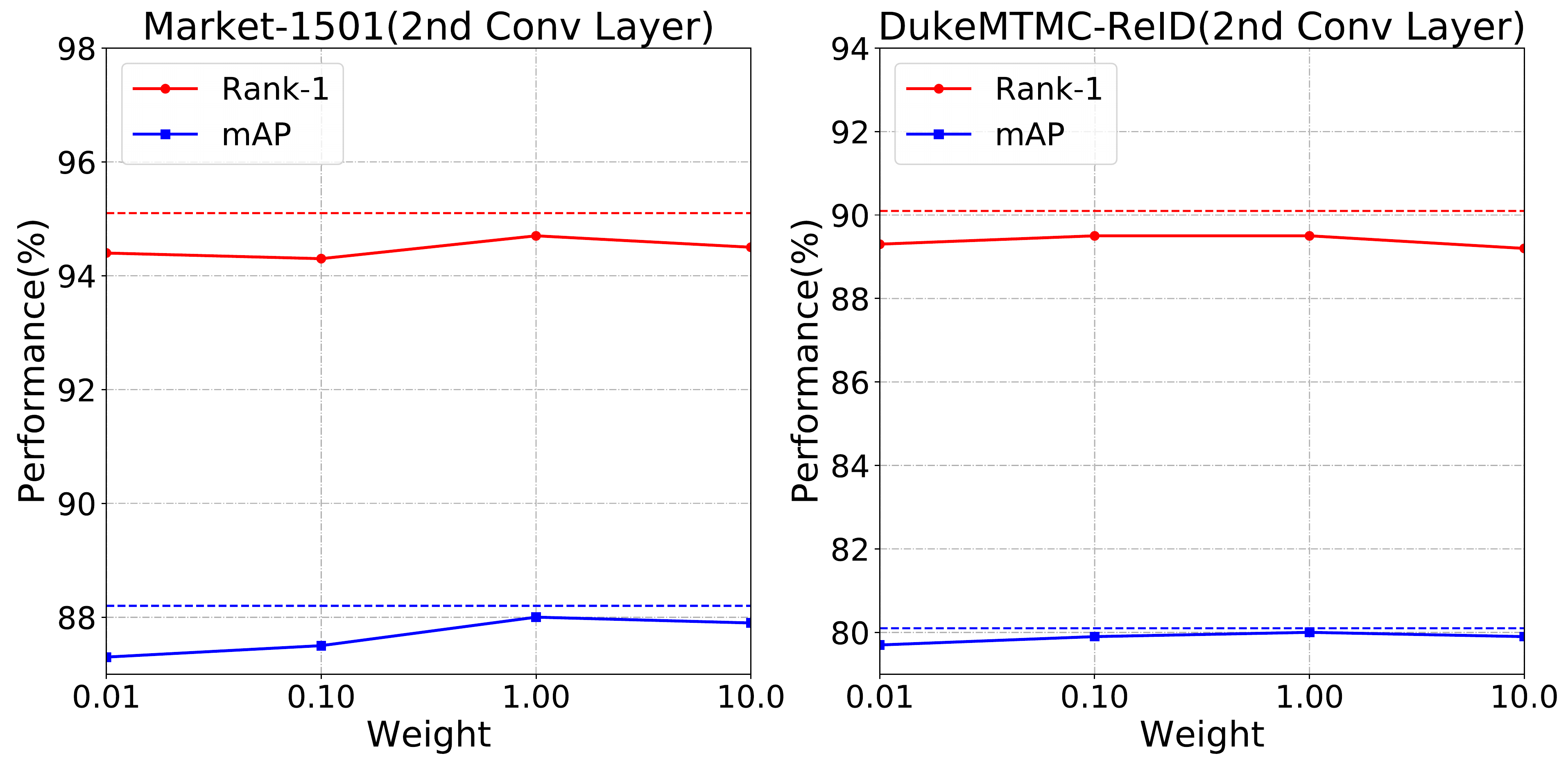}
\end{minipage}
}
\caption{
Performance comparison between the hard and soft parameter sharing strategies.
The horizontal axis stands for the weight of the $L2$ loss.
The red and blue dashed lines represent the Rank-1 accuracy and mAP via hard parameter sharing, respectively.
The solid lines denote the performance of soft parameter sharing.
(a) Experiments on the first $1\times1$ Conv layers of MTs and ATs.
(b) Experiments on the second $1\times1$ Conv layers of MTs and ATs.
(Best viewed in color.)}
\label{fig:hardVsSoft}
\end{figure}

Results of the comparison are illustrated in Fig.~\ref{fig:hardVsSoft}.
It is shown that hard parameter sharing consistently outperforms soft parameter sharing on each of the two $1\times1$ Conv layers.
Taking the experiments on the first $1\times1$ Conv layer as an example, hard parameter sharing outperforms the best performance of soft parameter sharing
by 0.4\%/0.3\% on Market-1501 and 0.3\%/0.4\% on DukeMTMC-ReID in terms of Rank-1 accuracy and mAP, respectively.
The above experiments demonstrate the effectiveness of hard parameter sharing for PSA in MPN.

\begin{table}
\caption{Performance Comparison with Variants for FSA (without PSA)}
\centering
\begin{tabular}{c|cc|cc}
\hline
  Dataset & \multicolumn{2}{c|}{Market-1501} & \multicolumn{2}{c}{DukeMTMC-ReID} \\
  \hline
  Metric                                 &Rank-1 &mAP    &Rank-1 &mAP \\
  \hline
  \hline
  Baseline                               &94.2   &84.4   &88.2   &77.4 \\
  Na\"ive MTL                              &94.6   &87.2   &88.7   &78.8 \\
  \hline
  Batch-wise                             &94.7   &87.8   &89.7   &80.2 \\
  Sample-wise~\cite{8zhang2018densely}   &95.1   &87.8   &90.1   &80.6 \\
  Sample-wise (Eq.~\ref{eq:SampleDist})  &95.0   &87.9   &90.1   &80.5 \\
  \hline
  Class-wise                             &95.5   &88.3   &90.5   &80.7 \\
  \hline
\end{tabular}
\label{tab:feaAlign}
\end{table}

\begin{table}
\caption{Performance Comparison with Variants for FSA (with PSA)}
\centering
\begin{tabular}{c|cc|cc}
\hline
  Dataset & \multicolumn{2}{c|}{Market-1501} & \multicolumn{2}{c}{DukeMTMC-ReID} \\
  \hline
  Metric                                 &Rank-1 &mAP    &Rank-1 &mAP \\
  \hline
  \hline
  Batch-wise                             &95.4   &88.1   &90.1   &80.3 \\
  Sample-wise~\cite{8zhang2018densely}   &95.7   &88.6   &90.6   &81.2 \\
  Sample-wise (Eq.~\ref{eq:SampleDist})  &95.6   &88.7   &90.5   &81.2 \\
  \hline
  Class-wise                             &96.1   &89.2   &91.2   &81.6 \\
  \hline
\end{tabular}
\label{tab:feaAlignWithPSA}
\end{table}

\subsubsection{Comparisons with Variants for FSA}
The next step is to compare the proposed class-wise FSA method with some possible variants.
Three variants are considered: batch-wise constraint, sample-wise constraint in Eq.~\ref{eq:SampleDist}, and sample-wise constraint in~\cite{8zhang2018densely}.
Each type of constraint is applied to the concatenated features of $K$ body parts, respectively.
The batch-wise constraint penalizes the cosine distance of the mean representations of the complete batch between MTs and ATs, ignoring the label information;
this is similar to the popular Maximum Mean Discrepancies (MMD) approach~\cite{long2015learning} to domain adaptation.
The second type of sample-wise constraint is realized according to the descriptions in~\cite{8zhang2018densely}:
in brief, we sum the features $\bf{h}$ and $\bf{g}$ in an element-wise manner for each image, then apply the triplet loss to the summed features rather than to $\bf{h}$ only.

Results of the comparison are tabulated in Table~\ref{tab:feaAlign} and Table~\ref{tab:feaAlignWithPSA}. We can make the following observations.
First, all four types of constraints can promote the performance of the na\"ive MTL model that adopts no constraint between MTs and ATs;
this indicates that alignment in the feature space can robustly promote the quality of the representations of MTs.
Second, the batch-wise constraint is inferior to both the sample-wise constraints and the proposed class-wise constraint.
This is because the batch-wise constraint is not discriminative, as it neglects the labels of the samples.
Third, the proposed class-wise constraint achieves the best performance. In particular, it outperforms both sample-wise constraints.
We can thus speculate that the quality of the representations of ATs is limited due to the errors in the prior of the body part locations.
Therefore, sample-wise constraints are rigid.
By contrast, the class-wise constraint is more robust to these errors via the averaging operation on samples for each class,
with the result that it achieves the best performance in both Table~\ref{tab:feaAlign} and Table~\ref{tab:feaAlignWithPSA}.

\subsection{Comparisons with State-of-the-Art Methods}
We compare the performance of MPN with state-of-the-art methods on four large-scale benchmark datasets:
Market-1501~\cite{27zheng2015scalable}, DukeMTMC-ReID~\cite{59zheng2017unlabeled}, CUHK03~\cite{33li2014deepreid}, and MSMT17~\cite{80PTGAN}.
According to the properties of the features, methods in this subsection are divided into three groups:
holistic feature-based methods, single-scale part feature-based methods, and multi-scale feature-based methods.
In the following, these are abbreviated as HF-, SPF-, and MSF-based methods, respectively.
The proposed MPN model belongs to the category of SPF-based methods.

\subsubsection{Performance Comparisons on Market-1501}
Comparison results are tabulated in Table~\ref{tab:1501Compare}. The following observations can be made.
First, MPN outperforms all state-of-the-art approaches in terms of both Rank-1 accuracy and mAP.
In particular, with the same backbone model (i.e. ResNet-50), MPN outperforms the DSA-Local(Single)~\cite{8zhang2018densely} approach by 2.3\% (96.3\%-94.0\%) in terms of Rank-1 accuracy and 6.2\% (89.4\%-83.2\%) in terms of mAP under the single-query mode.
Moreover, there are another two important advantages of MPN: 1) its model in the training stage is much more compact than that of DSA;
2) MPN requires only coarse priors, which are easy to obtain, while DSA depends on fine-grained 3D priors.
This comparison justifies the effectiveness of the parameter and feature space alignment strategies utilized in MPN.

Second, with single-scale part features, MPN outperforms all existing MSF-based approaches~\cite{60fu2018horizontal,2wang2018learning,8zhang2018densely,102zheng2019pyramidal}.
Multi-scale features are usually adopted to mitigate the problem of body part misalignment~\cite{60fu2018horizontal,2wang2018learning,102zheng2019pyramidal}.
This comparison indicates that features extracted via MPN have been semantically well-aligned and are therefore powerful.

Third, some recent HF-based methods also achieve competitive performance by enhancing the representation power of the backbone models~\cite{85RAGA,63ahmed2019Omni,86IAN,dai2019batch}.
For example, IANet~\cite{86IAN} and RGA-SC~\cite{85RAGA} insert attention modules into the backbone model that highlight body-relevant information.
The contributions of IANet~\cite{86IAN} and RGA-SC~\cite{85RAGA} are complementary to that in this paper.
Therefore, we also equip the backbone of MPN with the spatial attention module in~\cite{85RAGA}.
Hyper-parameters of the attention module are kept the same as in the original paper.
The combined model is denoted as MPN$^*$ in Table~\ref{tab:1501Compare}.
It can be seen that the performance of MPN is further promoted.

\begin{table}
\caption{Performance Comparisons on Market-1501}
\centering
\begin{tabular}{c|c|cc|cc}
\hline
\multicolumn{2}{c|}{\multirow{2}{*}{Methods}} & \multicolumn{2}{c|}{Single Query} & \multicolumn{2}{c}{Multiple Query} \\ \cline{3-6}
  \multicolumn{2}{c|}{} &Rank-1  & mAP  & Rank-1 & mAP \\
\hline
\hline
 \multirow{13}*{\rotatebox{90}{HF-based}}
  &HGD~\cite{matsukawa2019hierarchical} &87.0 &70.9 &- &- \\
  &PSE \cite{saquib2018pose} &87.7 &69.0 &- &- \\
  &SNL~\cite{li2018support} &88.3 &73.4 &92.1 &80.3 \\
  &DaRe \cite{62wang2018resource} &89.0 &76.0 &- &- \\
  &MLFN \cite{104chang2018multi} &90.0 &74.3 &92.3 &82.4 \\
  &Mancus \cite{34macus2018} & 93.1 &82.3 & 95.4 & 87.5 \\
  &SFT~\cite{luo2019spectral} & 93.4 &82.7 & - & - \\
  &DNN+CRF \cite{103chen2018group} &93.5 &81.6 & - & - \\
  &PGR~\cite{li2019pose} &93.9 &77.2 & - & - \\
  &IANet~\cite{86IAN} &94.4 &83.1 & - & - \\
  &OSNet~\cite{63ahmed2019Omni} &94.8 &84.9 &- & - \\
  &DCDS~\cite{alemu2019deep} &94.8 &85.8 & - & - \\
  &BDB+Cut~\cite{dai2019batch} &95.3 &86.7 &- & - \\
\hline\hline
 \multirow{11}*{\rotatebox{90}{SPF-based}}
  &PAR \cite{40zhao2017deeply} &81.0 &63.4 & - & - \\
  &AACN \cite{29xu2018attention} & 85.9 &66.9 & 89.8 &75.1 \\
  &Part-Aligned \cite{4suh2018part} & 91.7 & 79.6 & 94.0 & 85.2 \\
  &PCB~\cite{5sun2017beyond} & 92.3 &77.4 & - & - \\
  &PCB+RPP \cite{sun2019learning} & 93.8 &81.6 & - & - \\
  &DSA-Local(Single) \cite{8zhang2018densely} &94.0 &83.2 & - & - \\
  &FANN~\cite{zhou2019discriminative} &94.4 &82.5 & - & - \\
  &Auto-ReID~\cite{quan2019auto} &94.5 &85.1 & - & - \\
  &MHN-6 (PCB)~\cite{chen2019mixed} &95.1 &85.0 & - & - \\
  &\bfseries MPN &{\bfseries 96.3} &{\bfseries 89.4} &{\bfseries 97.0} &{\bfseries 92.7} \\
  &\bfseries MPN$^*$ &{\bfseries 96.4} &{\bfseries 90.1} &{\bfseries 97.3} &{\bfseries 93.1} \\
\hline\hline
 \multirow{8}*{\rotatebox{90}{MSF-based}}
  &PL-NET \cite{6yao2017deep} &88.2 &69.3 & & \\
  &HA-CNN \cite{38li2018harmonious} &91.2 &75.7 &93.8 &82.8 \\
  &HPM \cite{60fu2018horizontal} &94.2 &82.7  & - & - \\
  &MuDeep~\cite{qian2019leader} &95.3 &84.7 &- &- \\
  &FPR~\cite{he2019foreground}  &95.4 &86.6 &- &- \\
  &MGN \cite{2wang2018learning} &95.7 &86.9 &96.9 &90.7 \\
  &DSA-reID \cite{8zhang2018densely} & 95.7 &87.6 & - & - \\
  &Pyramid~\cite{102zheng2019pyramidal} &95.7 &88.2 & - & - \\
\hline
\end{tabular}
\label{tab:1501Compare}
\end{table}

\subsubsection{Performance Comparisons on DukeMTMC-ReID}
Comparison results on the DukeMTMC-ReID database are summarized in Table~\ref{tab:dukeCompare}.
From the table, it can be seen that MPN outperforms all other SPF-based methods by significant margins.
For example, MPN beats the PCB+RPP method~\cite{sun2019learning} that adopt the same backbone model (i.e. ResNet-50) by 7.0\% (91.5\%-84.5\%)
in terms of Rank-1 accuracy and 10.5\% (82.0\%-71.5\%) in terms of mAP.
It also outperforms one of the most recent HF-based methods, i.e. BDB+Cut~\cite{dai2019batch}, by 2.5\% and 6.0\% respectively in terms of Rank-1 accuracy and mAP.
Moreover, even when compared with one complex MSF-based method~\cite{102zheng2019pyramidal},
MPN still achieves a large performance improvement margin as high as 2.5\% and 3.0\% in terms of Rank-1 accuracy and mAP, respectively.

The above comparison results are consistent with those obtained on the Market-1501 database.
These experimental results justify the proposed methods' effectiveness at solving the body part misalignment problem for ReID.

\begin{table}
\centering\caption{Performance Comparisons on DukeMTMC-ReID}
\centering
\begin{tabular}{c|c|cc}
\hline
  \multicolumn{2}{c|}{Methods} & Rank-1 & mAP \\
\hline
\hline
 \multirow{12}*{\rotatebox{90}{HF-based}}
  & BraidNet~\cite{87BradNet} &76.4 &59.5 \\
  & SVDNet~\cite{56sun2017svdnet} &76.7 &56.8\\
  & PSE \cite{saquib2018pose} &79.8 &62.0 \\
  & GSRW~\cite{shen2018deep} &80.7 &66.4 \\
  & DuATM~\cite{si2018dual} &81.8 &64.6 \\
  & PGR~\cite{li2019pose} &83.6 &66.0 \\
  & DNN+CRF \cite{103chen2018group} &84.9 &69.5 \\
  & Mancus~\cite{34macus2018} & 84.9 &71.8 \\
  & SFT~\cite{luo2019spectral} &86.9 &73.2 \\
  & IANet~\cite{86IAN} &87.1 &73.4 \\
  & OSNet~\cite{63ahmed2019Omni} &88.6 &73.5 \\
  & BDB+Cut~\cite{dai2019batch} &89.0 &76.0 \\
\hline\hline
  \multirow{7}*{\rotatebox{90}{SPF-based}}
  & AACN \cite{29xu2018attention} &76.8 &59.3\\
  & PCB~\cite{5sun2017beyond} &81.8 &66.1 \\
  & Part-aligned \cite{4suh2018part} &84.4 &69.3 \\
  & PCB+RPP~\cite{sun2019learning} &84.5 &71.5 \\
  & FANN~\cite{zhou2019discriminative} &85.2 &70.2 \\
  & MHN-6 (PCB)~\cite{chen2019mixed} &89.1 &77.2 \\
  &\bfseries MPN &{\bfseries 91.5} &{\bfseries 82.0} \\
\hline\hline
  \multirow{7}*{\rotatebox{90}{MSF-based}}
  & HA-CNN \cite{38li2018harmonious} &80.5 &63.8\\
  & DSA-reID \cite{8zhang2018densely} &86.2 &74.3 \\
  & HPM \cite{60fu2018horizontal} &86.6 &74.3 \\
  & MuDeep~\cite{qian2019leader} &88.2 &75.6 \\
  & FPR~\cite{he2019foreground}  &88.6 &78.4 \\
  & MGN \cite{2wang2018learning} &88.7 &78.4 \\
  & Pyramid~\cite{102zheng2019pyramidal} &89.0 &79.0 \\
\hline
\end{tabular}
\label{tab:dukeCompare}
\end{table}

\subsubsection{Performance Comparisons on CUHK03}
We next compare the performance of MPN with that of the state-of-the-art approaches on the CUHK03 database.
Results of this comparison are presented in Table~\ref{tab:cuhk03Compare}.
Both manually labelled and auto-detected bounding boxes are employed for evaluation.

Results show that MPN still outperforms all other methods in Table~\ref{tab:cuhk03Compare} by large margins.
In particular, it outperforms the PCB+RPP approach~\cite{sun2019learning}, which is based on the same backbone model, by 19.7\% in terms of Rank-1 accuracy and 21.6\% in terms of mAP.
Note that another advantage of MPN relative to PCB+RPP is that MPN can be trained with the standard end-to-end strategy in a single stage;
by contrast, PCB+RPP depends on a four-stage training scheme, as its PCB and RPP modules have to be optimized sequentially~\cite{sun2019learning}.
MPN also outperforms another two most recent SPF-based approaches~\cite{chen2019mixed,quan2019auto} that adopt more powerful backbone models.
Furthermore, when compared with one of the most recent MSF-based methods (i.e. Pyramid~\cite{102zheng2019pyramidal}), MPN still exhibits a clear advantage.
In brief, its Rank-1 accuracy is higher than that of Pyramid by 4.5\% and 6.1\% on CUHK03-Detected and CUHK03-Labeled data, respectively.
The above comparisons justify the effectiveness of MPN.

\begin{table}
\centering
\caption{Performance Comparisons on CUHK03}
\begin{tabular}{c|c|cc|cc}
\hline
  \multicolumn{2}{c|}{\multirow{2}{*}{Methods}} & \multicolumn{2}{c|}{Detected} & \multicolumn{2}{c}{Labeled} \\ \cline{3-6}
  \multicolumn{2}{c|}{} &Rank-1  & mAP  & Rank-1 & mAP \\
\hline
\hline
  \multirow{8}*{\rotatebox{90}{HF-based}}
  & PAN \cite{57zheng2018pedestrian} &36.3 &34.0 &36.9 &35.0 \\
  & SVDNet \cite{56sun2017svdnet} &41.5 &37.3 & - & - \\
  & MGCAM \cite{12song2018mask} &46.7 & 46.9 &50.1 &50.2 \\
  & Rolling-back~\cite{ro2019backbone} &55.6 &50.5 &59.8 &55.7 \\
  & SFT~\cite{luo2019spectral} & & &68.2 &62.4 \\
  & Mancus \cite{34macus2018} & 65.5 &60.5 &69.0 &63.9 \\
  & OSNet \cite{63ahmed2019Omni} &72.3 &67.8 &- &- \\
  & BDB+Cut~\cite{dai2019batch} &76.4 &73.5 &79.4 &76.7\\
\hline\hline
  \multirow{6}*{\rotatebox{90}{SPF-based}}
  & PCB \cite{5sun2017beyond} &61.3 &54.2 & - & - \\
  & PCB+RPP \cite{sun2019learning} &63.7 &57.5 & - & - \\
  & HPDN \cite{zhang2018person} &- &- &64.3 &58.2\\
  & MHN-6 (PCB)~\cite{chen2019mixed} &71.7 &65.4 &77.2 &72.4 \\
  & Auto-ReID~\cite{quan2019auto} &73.3 &69.3 &77.9 &73.0 \\
  & \bfseries MPN &{\bfseries 83.4} &{\bfseries 79.1} &{\bfseries 85.0} &{\bfseries 81.1} \\
\hline\hline
 \multirow{6}*{\rotatebox{90}{MSF-based}}
  & HA-CNN \cite{38li2018harmonious} &41.7 &38.6 &44.4 &41.0 \\
  & HPM \cite{60fu2018horizontal} &63.9 &57.5 & - & - \\
  & MGN \cite{2wang2018learning} &66.8 &66.0 &68.0 &67.4 \\
  & MuDeep~\cite{qian2019leader} &71.9 &67.2 &75.6 &70.5 \\
  & DSA-reID \cite{8zhang2018densely} &78.2 &73.1 &78.9 &75.2  \\
  & Pyramid~\cite{102zheng2019pyramidal} &78.9 &74.8 &78.9 &76.9 \\
\hline
\end{tabular}
\label{tab:cuhk03Compare}
\end{table}

\subsubsection{Performance Comparisons on MSMT17}
Finally, we evaluate the performance of MPN on the MSMT17 database, which features complex background and illumination changes.
As MSMT17 was released only relatively recently, only a few works have conducted experiments on this database.
We compare the performance of MPN with these methods in Table~\ref{tab:msmt17Compare}.
In this table, we merge the SPF- and MSF-based methods into one category, which is named `Part-based methods'.

Experimental results demonstrate that MPN outperforms all other methods by significant margins,
which is consistent with the experimental results on the first three databases.
For example, MPN outperforms one of the most recent methods, i.e. OSNet~\cite{63ahmed2019Omni}, by 4.8\% and 9.8\% in terms of Rank-1 accuracy and mAP, respectively.
The above comparisons justify the effectiveness of MPN for pedestrian images in more complex scenes.

\begin{table}
\centering\caption{Performance Comparisons on MSMT17}
\centering
\begin{tabular}{c|c|cc}
\hline
  \multicolumn{2}{c|}{Methods} & Rank-1 & mAP \\
\hline
\hline
  \multirow{6}*{\rotatebox{90}{HF-based}}
  & Verif-Identif \cite{zheng2018discriminatively, 78dgnet} &60.5 &31.6  \\
  & PGR~\cite{li2019pose} &66.0 &37.9 \\
  & SFT~\cite{luo2019spectral} &73.6 &47.6 \\
  & IANet\cite{86IAN} &75.5 &46.8 \\
  & DG-Net~\cite{78dgnet} &77.2 &52.3 \\
  & OSNet \cite{63ahmed2019Omni} & 78.7 & 52.9 \\
\hline\hline
  \multirow{6}*{\rotatebox{90}{Part-based}}
  & PDC \cite{11su2017pose, 80PTGAN} & 58.0 &29.7  \\
  & GLAD \cite{wei2017glad, 80PTGAN} &61.4 &34.0 \\
  & PCB+RPP~\cite{sun2019learning} &69.8 &43.6 \\
  & Our Baseline                   &72.4 &47.5 \\
  & Auto-ReID~\cite{quan2019auto} &78.2 &52.5 \\
  & \bfseries MPN &{\bfseries 83.5} &{\bfseries 62.7} \\
\hline
\end{tabular}
\label{tab:msmt17Compare}
\end{table}

\subsubsection{Comparisons of Model Complexity}
In this experiment, we demonstrate that MPN not only achieves superior performance in terms of ReID accuracy,
but also offers advantages in terms of both its time and space complexities.
Four powerful part-based approaches are compared: PCB~\cite{5sun2017beyond}, MGN~\cite{2wang2018learning}, Pyramid~\cite{102zheng2019pyramidal}, and DSA-reID~\cite{8zhang2018densely}.
All four of these models adopt the ResNet-50 backbone model, meaning that they are directly comparable.
To further facilitate fair comparison, input images for all the five models are resized to $384\times128$ pixels.
Batch sizes of all methods are also unified.
Moreover, the number of parameters for classification layers in the training stage depends on the identity number of each database;
therefore, their parameters are not taken into account for all models.

Since the CA modules are optional for MPN, we here test the model complexity of MPN$^o$.
It is worth noting that MPN$^o$ also consistently outperforms all other models in terms of ReID accuracy, as shown in Table~\ref{tab:ablation}.
Comparisons are conducted on a Titan V GPU, and results are summarized in Table~\ref{tab:eff}.
The time cost in Table~\ref{tab:eff} refers to the average time required to process one image.

We can thus make the following observations. First, the time cost of MPN$^o$ is only slightly higher than that of a very basic part-based model, i.e. PCB~\cite{5sun2017beyond}.
Moreover, the time cost of MPN$^o$ is lower than that of~\cite{2wang2018learning,102zheng2019pyramidal,8zhang2018densely} in both the training and testing stages.
Second, except for the classification layers, the number of parameters for MPN$^o$ is the same at both the training and testing stages;
by contrast, the model size of DSA-reID~\cite{8zhang2018densely} in the training stage is significantly larger than that in the testing stage.
Third, compared with~\cite{5sun2017beyond,102zheng2019pyramidal} (the models of which are also compact in the training stage),
MPN$^o$ solves the body part misalignment problem more effectively.
In fact, MPN$^o$ has more parameters than~\cite{5sun2017beyond,102zheng2019pyramidal} as it adopts more $1\times1$ Conv layers,
which are computationally very efficient in practice.

Accordingly, the above comparisons demonstrate that the proposed MPN model is both compact and efficient.

\begin{table}
\centering
\caption{Comparisons of Model Complexity}
\centering
\begin{tabular}{c|cc|cc}
\hline
\multirow{2}*{\tabincell{c}{Methods}} &\multicolumn{2}{c|}{Training} &\multicolumn{2}{c}{Testing} \\ \cline{2-5}
          &\# params &time cost &\# params &time cost \\
\hline
PCB~\cite{5sun2017beyond}             &26.8M   &8.3ms   &26.8M     &4.9ms    \\
MGN~\cite{2wang2018learning}          &68.8M   &16.2ms  &68.8M     &11.9ms \\
Pyramid~\cite{102zheng2019pyramidal}  &29.1M   &14.5ms  &29.1M     &6.4ms  \\
DSA-reID~\cite{8zhang2018densely}     &187.8M  &34.7ms  &38.5M     &5.4ms  \\
\hline
MPN$^o$                               &31.5M   &10.9ms  &31.5M     &5.1ms  \\
\hline
\end{tabular}
\label{tab:eff}
\end{table}

\section{Conclusion}
In this paper, we propose a robust, compact, and easy-to-use model, named Multi-task Part-aware Network (MPN), to extract semantically aligned part-level representations.
In the training stage, MPN includes one main task (MT) and one auxiliary task (AT) for each body part.
We equip ATs with a coarse prior of body part locations for training images, and further propose a dual alignment mechanism, i.e. parameter and feature space alignments,
to guide the MTs in learning high-quality parameters for part-level feature extraction.
In the testing stage, the ATs are removed, and only MTs are saved for feature extraction;
therefore, MPN is freed from body part detection during inference.
Due to the innovations of our design, the time and space complexities of MPN are only slightly increased relative to a very basic part-based model~\cite{5sun2017beyond} at both the training and testing stages.
At the training stage, MPN is also robust to coarse priors, which are very easy to obtain.
Moreover, comparisons on four large-scale ReID databases demonstrate that MPN significantly outperforms existing approaches at a relatively small computational cost.
Therefore, MPN can be surmised to be both powerful and easily applicable to practical ReID applications.



\ifCLASSOPTIONcompsoc



\bibliographystyle{IEEEtran}
\bibliography{egbib}
\end{document}